\documentclass[conference]{IEEEtran}
\IEEEoverridecommandlockouts
\usepackage{amsmath,amssymb,amsfonts}
\usepackage{algorithmic}
\usepackage{graphicx}
\usepackage{textcomp}
\PassOptionsToPackage{square,numbers,sort&compress}{natbib}
\usepackage[colorlinks=true, bookmarksopen,
			linkcolor={lirio_blue},
			anchorcolor={black},
			citecolor={lirio_blue},
			filecolor={magenta},
			menucolor={black},
			plainpages=false,pdfpagelabels,
			urlcolor={lirio_blue}]{hyperref}
\usepackage{caption}
\usepackage{amsfonts, amsmath, amssymb, amsthm}
\usepackage{bbm}
\usepackage{bm}
\usepackage{nicefrac}
\usepackage{microtype}
\usepackage{booktabs}
\usepackage[dvipsnames]{xcolor}
\usepackage{subcaption}

\newcommand{\start}{\begin{equation}}
\newcommand{\finish}{\end{equation}}
\newcommand{\s}{\begin{displaymath}}
\newcommand{\f}{\end{displaymath}}

\newcommand{\sa}{\[\begin{aligned}}
\newcommand{\fa}{\end{aligned}\]}
\newcommand{\sbmatrix}{\begin{bmatrix}}
\newcommand{\fbmatrix}{\end{bmatrix}}
\renewcommand{\baselinestretch}{1}
\newcommand{\R}{{\mathbb R}}

\newcommand{\E}{{\mathbb E}}

\definecolor{lirio_orange}{HTML}{f7800a}
\definecolor{lirio_gray}{HTML}{686868}
\definecolor{lirio_blue}{HTML}{0A4150}
\definecolor{lirio_Borange}{HTML}{F59105}

\begin{document}

\title{
% Improving policy entropy for personalization tasks in policy gradient agents with MMD regularization
% MMD regularization improves policy entropy in policy gradient agents on personalization tasks
%MMD-based regularization for improving entropy in policy gradient agents on personalization tasks
Increasing Entropy to Boost Policy Gradient Performance on Personalization Tasks
%regularization for improving performance in policy gradient agents on personalization tasks
% MMD-PG: improving policy entropy for personalize
% Diversity-promoting policy regularization for personalized reinforcement learning agents\\
}

\author{\IEEEauthorblockN{Andrew~Starnes}
\IEEEauthorblockA{\textit{Lirio AI Research, Lirio LLC.} \\
Knoxville, U.S.A. \\
astarnes@lirio.com}
\and
\IEEEauthorblockN{Anton Dereventsov}
\IEEEauthorblockA{\textit{Lirio AI Research, Lirio LLC.}\\
Knoxville, U.S.A. \\
adereventsov@lirio.com}
\and
\IEEEauthorblockN{ Clayton~Webster}
\IEEEauthorblockA{\textit{Lirio AI Research, Lirio LLC.}\\
Knoxville, U.S.A. \\
cwebster@lirio.com}
}

\maketitle

%%%%%=====================
\begin{abstract}
In this effort, we consider the impact of regularization on the diversity of actions taken by policies generated from reinforcement learning agents trained using a policy gradient.
Policy gradient agents are prone to entropy collapse, which means certain actions are seldomly, if ever, selected. 
We augment the optimization objective function for the policy with terms constructed from various $\varphi$-divergences and Maximum Mean Discrepancy which encourages current policies to follow different state visitation and/or action choice distribution than previously computed policies.
We provide numerical experiments using MNIST, CIFAR10, and Spotify datasets.
The results demonstrate the advantage of diversity-promoting policy regularization and that its use on gradient-based approaches have significantly improved performance on a variety of personalization tasks.
Furthermore, numerical evidence is given to show that policy regularization increases performance without losing accuracy.
\end{abstract}
%%%%%=====================


%%%%%=====================
\section{Introduction}
\label{sec:intro}
%%%%%=====================
\noindent
Recommendation system models (see, e.g.,~\cite{10.1007/978-1-4899-7637-6_1,10.1609/aimag.v32i3.2361})
have become critically important in retaining customers of industries such as retail, e-commerce, media apps, or even healthcare.
Corporations like Netflix, Spotify, and Amazon, use sophisticated collaborative filtering and content-based recommendation systems for video, song, and/or product recommendations~\cite{10.1145/2843948,10.1007/978-1-4899-7637-6_11,10.1145/2959100.2959120,10.1109/MIC.2017.72,10.1145/2623372}.
For a recent overview of recommender systems in the healthcare domain see, e.g.,~\cite{10.1007/s10844-020-00633-6} and the references therein.

Conventional personalization focuses on personal, transactional, demographic, and possibly health-related information, such as an individual's age, residential location, employment, purchases, medical history, etc. Additional applications of personalization include: web content personalization and layout customization~\cite{10.1016/j.jss.2016.02.008, Ricci2011IntroductionTR}; customer-centric interaction with healthcare providers~\cite{lasalvia2020personalization, 10.1145/3318236.3318249, lei2017actor, zhu2018robust, hassouni2018personalization, tan2020adaptive}; personalized medical treatments~\cite{2007_aspinall, 10.1016/S0167-7799(01)01814-5}. One of the major challenges associated with personalization techniques is the time required to adapt and update such approaches to changes in individual behaviors, reactions, and choices.

Recently, reinforcement learning (RL) has been increasingly exploited in personalized recommendation systems that continually interact with users (see, e.g., \cite{10.3233/DS-200028} and the references therein).
As opposed to traditional recommendation techniques, RL is a more complex and transformative approach that considers behavioral and  real-time data produced as the result of user action.
Examples of this technique include online browsing behavior, communication history, in-app choices, and other engagement data.
This allows for more individualized experiences like adding personalized engaging sections to the body of an email or sending push notifications at a time when the customer is typically active, which results in more customized communication and thus, ultimately, greater conversion.

One of the major challenges associated with personalized RL agents is that standard optimization techniques often stall or even fail to converge when applied to such complex problems.  This results in highly localized policies having lower entropy which directly translates into very few actions taken by the agent throughout the training process.  Improving the diversity of actions taken by the policy is critical to improvining the performance of the RL agent on a variety personalization tasks~\cite{Anton:2021A}.

The traditional approach for combating low-entropy models is to regularize the standard objective with an entropy (penalty) term, such that the optimal policy additionally aims to maximize its entropy at each visited state, see, e.g.,~\cite{DBLP:journals/corr/abs-1812-05905, pmlr-v80-haarnoja18b, pmlr-v70-haarnoja17a} and the references therein.  This is achieved by subtracting a weighted term for the entropy of the model’s prediction from the loss function, thereby encouraging a more entropic model. This is equivalent to adding the  Kullback-Leibler (KL) divergence between the policy and the uniform distribution.

Comparing probability distributions is a fundamental component of many 
supervised, unsupervised, and RL problems.
In the machine learning community, the first discrepancies that were introduced to compare two probability distributions are $\varphi$-divergences \cite{10.1214/aop/1176996454}, with $\varphi$ is a convex, lower semi-continuous function such that $\varphi(1)=0$.  Such divergences can be viewed as a weighted average (by $\varphi$) of the odds-ratio between the two measures. In particular, we compute the following
\start
    D_{\varphi}(\alpha\|\beta)
    = \mathbb{E}_{\beta} \bigg(\varphi\Big(\frac{\alpha}{\beta}\Big)\bigg).
\finish
The computational simplicity of $\varphi$-divergences has made them very popular; with the most widely used being the KL divergence (see, e.g., Table~\ref{fig:phi-divergences} and the work \cite[Section 2]{genevay:tel-02319318}).

% Table I
\begin{table*}[t]
    \centering
    % \caption{Definitions of $\varphi$-divergences ($\mathcal{M}(\mathcal{A})$ denotes the $\alpha$ and $\beta$-measurable sets of $\mathcal{A}$)}
    \caption{Definitions of $\varphi$-divergences.}
    \label{fig:phi-divergences}
    \begin{tabular}{|c|c|c|}
        \hline
         Kullback-Leibler
            &$D_{\text{KL}}(\alpha\|\beta)
                % =\int_{\mathcal{A}}\alpha(a)\ln\left(\frac{\alpha(a)}{\beta(a)}\right)\;da
                =\E_{\alpha}\left(\ln\left(\frac{\alpha(a)}{\beta(a)}\right)\right)
            $
            &$\varphi(x)=x\ln x$\\
        \hline
        Jensen-Shannon
            &$D_{\text{JS}}(\alpha\|\beta)
                =D_{\text{KL}}(\alpha\|\tfrac{1}{2}(\alpha+\beta))+D_{\text{KL}}(\beta\|\tfrac{1}{2}(\alpha+\beta))
            $
            &$\varphi(x)=x\ln x-(1+x)\ln\left(\frac{1+x}{2}\right)$\\
        \hline
        Hellinger
            &$D_{H^2}(\alpha\|\beta)
                % =\int_{\mathcal{A}}\Big(\sqrt{\alpha(a)}-\sqrt{\beta(a)}\Big)^2\;da
                =\E_{\frac{1}{2}(\alpha+\beta)}\Big(\big(\sqrt{\alpha(a)}-\sqrt{\beta(a)}\big)^2\Big)
            $
            &$\varphi(x)=(\sqrt{x}-1)^2$\\
        \hline
        Total-Variation
            &$D_{TV}(\alpha\|\beta)
                =\sup_{A\in\mathcal{M}(\mathcal{A})}\Big|\alpha(A)-\beta(A)\Big|
            $\textsuperscript{\dag}
            &$\varphi(x)=\tfrac{1}{2}|x-1|$\\
        \hline
    \end{tabular}
    \\[2ex]
    \small\textsuperscript{\dag} $\mathcal{M}(\mathcal{A})$ denotes the $\alpha$ and $\beta$-measurable sets of $\mathcal{A}$
\end{table*}

However, such approaches suffer from the major drawback of not metrizing weak-convergence,
which is instrumental for discrepancies on measure, as it ensures that the losses remain stable under small perturbations of the support of the measures. A class of discrepancies that satisfy this requirement are known as Maximum Mean Discrepancies (MMD) \cite{NIPS2006_e9fb2eda}, which are a special case of integral probability metrics (IPM) \cite{10.2307/1428011}. Such approaches compare distributions without initially estimating their density functions.  MMD is defined by the notion of representing distances between distributions as distances between {\it mean embeddings} of features, where the feature map is a kernel from a reproducing kernel Hilbert space (RKHS). This family of discrepancies presents the advantage of being efficiently computed from samples~--- 
both statistically since the estimates are robust with a small number of samples (reduced complexity) and also numerically as it can be computed in closed form.

%\andrew{this needs to be updated to reflect our new focus}
In this work we augment the optimization objective function for the policy with various $\varphi$-divergence-based as well as MMD-based\footnote{MMD is the more popular IPM in machine learning applications, including, e.g., generative models (see \cite{Li2017MMDGT,Binkowski2018DemystifyingMG} and the references therin) due to the fact that it is applicable to a wide range of data types and distributions, computationally tractable even for high-dimensional data, and it is relatively robust to the curse of dimensionality \cite{genevay:tel-02319318}.}
term which encourages current policies to follow different state visitation and/or action choice distribution than previously computed policies. As such, by utilizing these more entropic variants of PG enables us to obtain a completely distinct set of policies.

Our main contributions are:
\begin{itemize}
    \item formalization of $\varphi$-divergence-based as well as MMD-based
    regularization for personalized tasks in contextual bandit problems; and 
    \item empirical demonstration of impact such regularization approaches  have on RL.
\end{itemize}

\subsection{Related work}
\label{sec:work}

The goal of this paper is to understand the impact that policy regularization has on an agent’s learning.
It is often observed that policy gradient algorithms suffer from premature convergence to semi-deterministic, suboptimal policies.
Avoiding this lack of diversity in actions is the motivation for adding entropy regularization to the REINFORCE algorithm \cite{williams1991function}, which is aptly named REINFORCE/MENT with MENT standing for Maximization of ENTropy.
Using entropy regularization has also been found to improve agent performance (e.g., \cite{mnih2016asynchronous}, \cite{ahmed2019understanding}).
While typical entropy regularization uses KL divergence between the policy and a uniform distribution over the actions, \cite{galashov2019information} uses KL divergence between the policy and the so-called default policy to improve performance.
Bergmann divergence is used in \cite{wang2019divergence} to more safely train on-policy agents with off-policy data.

The work \cite{NEURIPS2018_a2802cad} presents diversity-driven approach for exploration, which can be easily combined with both off- and on-policy reinforcement learning algorithms. The authors 
show that by simply adding a distance measure regularization to the loss function, the proposed methodology significantly enhances an agent’s exploratory behavior. Similarly, the effort \cite{10.24963/ijcai.2019/821} presents an MMD-based approach for identifying a collection of near-optimal policies with significantly different distributions of trajectories.  

Soft policy optimization was introduced in \cite{pmlr-v80-haarnoja18b, pmlr-v70-haarnoja17a}.
These works show that the impact of entropy regularization goes beyond providing the agent with extra exploration, 
but also serves as more stable training process by avoiding a collapse onto a select set of actions. Similar to our work, 
empirical results also show the robustness of these approaches when compared with standard optimization algorithms.

% Table II
\begin{table*}[t]
    \centering
    % \caption{$\varphi$-divergences up to an additive constant in a finite action space and their gradients ($a^*$ is the action maximizing $|\pi_{\theta}(a|s)-u(a)|$)}
    \caption{$\varphi$-divergences up to an additive constant in a finite action space and their gradients.}
    \label{fig:phi-divergences-finite-actions-space}
    \begin{tabular}{|c|c|c|c|}
        \hline
        \multicolumn{2}{|c|}{$\varphi$-Divergence}
            &Definition
            &Gradient (wrt $\theta$)\\
        \hline
         Kullback-Leibler
            &$D_{\text{KL}}(\pi_{\theta}(s)\|u)$
            &$\displaystyle\sum_{a=1}^{n}\pi_{\theta}(a\|s)\ln\pi_{\theta}(a\|s)$
            &$\displaystyle\sum_{a=1}^{n}\big(1+\ln\pi_{\theta}(a\|s)\big)\nabla\pi_{\theta}(a\|s)$\\
        \hline
        Jensen-Shannon
            &$D_{\text{JS}}(\pi_{\theta}(s)\|u)$
            % &$D_{\text{KL}}(\pi_{\theta}(s)\|\frac{1}{2}(\pi_{\theta}(s)+u))+D_{\text{KL}}(u\|\frac{1}{2}(\pi_{\theta}(s)+u))$
            &$D_{\text{KL}}(\pi_{\theta}(s)\|u)-D_{\text{KL}}(\frac{1}{2}(\pi_{\theta}(s)+u)\|u)$
            &$\displaystyle\sum_{a=1}^{n}\ln\left(\frac{2\pi_{\theta}(a|s)}{\pi_{\theta}(a|s)+u(a)}\right)\nabla\pi_{\theta}(a\|s)$\\
        \hline
        Hellinger
            &$D_{H^2}(\pi_{\theta}(s)\|u)$
            &$\displaystyle\sum_{a=1}^{n}\Big(\pi_{\theta}(a\|s)-2\sqrt{u(a)\pi_{\theta}(a\|s)}\Big)$
            &$\displaystyle\sum_{a=1}^{n}\left(1-\sqrt{\frac{u(a)}{\pi_{\theta}(a|s)}}\right)\nabla\pi_{\theta}(a\|s)$\\
        \hline
        Total-Variation
            &$D_{TV}(\pi_{\theta}(s)\|u)$
            &$\displaystyle\max_{a=1,...,n}\big|\pi_{\theta}(a|s)-u(a)\big|$
            &$\text{sgn}\big(\pi_{\theta}(a^*|s)-u(a^*)\big)\nabla\pi_{\theta}(a^*\|s)$\textsuperscript{\ddag}\\
        \hline
    \end{tabular}
    \\[2ex]
    \small\textsuperscript{\ddag} $a^*$ is the action maximizing $|\pi_{\theta}(a|s)-u(a)|$
\end{table*}

%%%%%=====================
\section{Background}
\label{sec:back}
%%%%%=====================
%
We consider a contextual bandit environment~\cite{langford2007epoch}, with a continuous state (context) space $\mathcal{S} \subset \mathbb{R}^m$, a discrete action space $\mathcal{A} = \{1, 2, \ldots, n\}$ consisting of $n$ available actions, and a reward function $r : \mathcal{S} \times \mathcal{A} \to \mathbb{R}$.  Using this conventional setting for recommendation and personalization tasks \cite{li2010contextual, tang2015personalized}, we define the reward $\mathcal{J}$ of the policy $\pi$, which is given by the expectation return under the policy, i.e.,
\start\label{eq:V}
    %\max_{{\bm\theta} \in \R^d} V(\pi_{\bm\theta}),
    %\text{ with }
    \mathcal{J}(\pi) = \mathbb{E} \Big[ r(s,a) \ \big|\ s \sim \mathcal{S}, a \sim \pi(s) \Big], 
\finish
where $\pi(s)$ denotes the action probability distribution as state $s$.
In this setting, traditional approaches for reinforcement learning aim to find a policy $\pi$ that maximizes the reward function $\mathcal{J}$.
However, to promote a more entropic model, we augment this optimization functional with various $\varphi$-divergence-based as well as an MMD-based regularization function $\mathcal{R}$.  In other words, without loss of generality, our goal is to find a policy $\pi$ that solves the following regularized optimization problem, namely:
\start\label{eq:RLopt}
\max_{{\theta} \in \R^d}\;\;
\mathcal{J}(\pi_{\theta}) + \lambda\,\mathcal{R}(\pi_{\theta}),
\finish
where ${\theta}=(\theta_1, \ldots, \theta_d)\in\R^d$ is a $d$-dimensional  parameter that represents, e.g., the weights of a neural network, and $\lambda\in\R$ is a regularization (penalty) parameter.  In what follows, we detail the construction of the regularized problem \eqref{eq:RLopt} and solutions will be sought for the policy gradient technique.

%%%%%=====================
\subsection{Relative entropy}
\label{sec:relative_entropy}
%%%%%=====================
\noindent
The distribution of the agent's policy $\pi$ is often critical in practical applications as it directly translates to the actions the agent is taking throughout the training process.
A conventional way to quantify the policy distribution is by computing its entropy $H(\pi)$ given by
\start
\label{eq:entropy}
    H(\pi) = \mathbb{E} \left[ \sum_{a \in \mathcal{A}} \pi(a|s) \log\pi(a|s) \ \big|\ s \sim \mathcal{S} \right].
\finish
Entropy quantifies the amount of uncertainty involved in the value of a random variable or the outcome of a random process.  In RL, entropy indicates how distributed the policy is, with more localized policies having lower entropy, which is known to lead to undesirable results, discussed in, e.g., \cite{dou2008evaluating}.

%%%%%=====================
\subsection{Policy gradient methods}
\label{sec:PGvQL}
%%%%%=====================
\noindent
Policy gradient (PG) makes use of gradients to iteratively optimize a policy $\pi_{\theta}(s,a)$, parameterized by $\theta\in\R^d$. 
% The standard objective function, denoted
% $\mathcal{J}_{\textsf{PG}}$, is the return $\mathcal{V}(\pi)$ given by \eqref{eq:V}.
In order to maximize $\mathcal{J}$, we apply the Policy Gradient Theorem (see 13.2 of \cite{sutton2018reinforcement} for example), which shows
\start
\begin{aligned}
% \label{eq:gradPG}
    % PG
    \nabla_\theta\mathcal{J}(\pi_{\theta})
    &=\sum_{a\in\mathcal{A}}r(s,a)\pi_{\theta}(a|s)\nabla\ln\pi_{\theta}(a|s)\\
    &=\mathbb{E}_{a\sim\pi_{\theta}(s)}(r(s,a)\nabla\ln\pi_{\theta}(a|s)).
\end{aligned}
\finish
We use a Monte Carlo approximation of this expectation in order to estimate the gradient, denoted $\nabla_\theta\mathcal{J}_{\textsf{PG}}$ and given by \eqref{eq:gradPG}.

%%%%%=====================
\section{Diversity promoting policy regularization}
\label{sec:dper}
%%%%%=====================

In this section we develop all the necessary machinery to improve the
existing policy gradient method by including an additional diversity-promoting term, thus, resulting in more entropic approaches.

%%%%%=====================
\subsection[divergence regularization]{$\varphi$-divergence regularization}
\label{sec:entropy_regularization}
%%%%%=====================

\noindent
The traditional approach for combating low-entropy models is to augment the standard objective with an entropy (penalty) term, such that the optimal policy additionally aims to maximize its entropy at each visited state \cite{pmlr-v70-haarnoja17a}. 
This is achieved by adding a weighted term that measures the diversity of the model's prediction from the loss function, thereby encouraging a more entropic model. 
One way to accomplish this is by adding any of the $\varphi$-divergences in Table~\ref{fig:phi-divergences} calculated between the policy $\pi_{\theta}$ and the uniform distribution. 
Table~\ref{fig:phi-divergences-finite-actions-space} provides simplified definitions of the $\varphi$-divergences in the case of a finite action space, where we use 
\begin{equation}\label{eq:uniform_policy}
    u\sim\text{Unif}(1,2,\ldots,n)
\end{equation}
as the uniform distribution here but can be replaced by any other distribution of the actions.
Therefore, using \eqref{eq:RLopt} and given a regularization constant $\lambda\in\R$, our goal is to find a policy $\pi$ that maximizes a new objective function, namely:
\start
\label{eq:KLopt}
\max_{{\theta} \in \R^d} \;\;
%\left[
\mathcal{J}(\pi_{\theta}) + 
\lambda\,D_{\varphi}(\pi_{\theta}\, ||\, u).
%\right].
\finish

%%%%%=====================
\subsection{MMD regularization}
\label{sec:MMD}
%%%%%=====================
\noindent 
In addition to $\varphi$-divergence regularization, defined in Table~\ref{fig:phi-divergences-finite-actions-space}, we also propose to exploit a family of discrepancies known as Maximum Mean Discrepancy (MMD). 
Given a RKHS $\mathcal{H}$ with kernel $k$, MMD between two probability measures $\alpha$ and $\beta$ is given by
\start 
\begin{aligned}
\label{eq:MMD}
&\textsf{MMD}_{k}^2(\alpha,\beta) : = 
\left(
\sup_{\{f: \|f\|_{\mathcal{H}} \leq 1\}}
\big |
\mathbb{E}_{\alpha} f(x) - \mathbb{E}_{\beta} f(y)
\big |
\right)^2 \\
& = \mathbb{E}_{\alpha\otimes\alpha} k(x,x') 
+ \mathbb{E}_{\beta\otimes\beta} k(y,y') 
-2 \mathbb{E}_{\alpha\otimes\beta} k(x,y).
\end{aligned}
\finish
This family of discrepancies presents the advantage of being efficiently estimated from samples of the measures~--- both statistically since the estimates are robust with a small number or samples (reduced complexity) and also numerically, as \eqref{eq:MMD} can be computed in closed form.
Therefore, using \eqref{eq:RLopt} and given a regularization constant $\lambda\in\R$, our goal is to find a policy $\pi$ that maximizes a new objective function, namely:
\start
\label{eq:MMDopt}
\max_{{\theta} \in \R^d} \;\;
%\left[
\mathcal{J}(\pi_{\theta}) + 
\lambda\,\textsf{MMD}_{k}^2(\pi_\theta,u).
%\right].
\finish

There are many choices for $k$ (or equivalently $\mathcal{H}$).
For our examples in this paper, we use the Gaussian kernel, $k(x,y)=\text{exp}(\|x-y\|^2)$, where $\|x-y\|=\mathbbm{1}(x=y)$ for $x,y\in\mathcal{A}$.
We do this because each of the examples focuses on correctly labeling and the arithmetic difference between two labels is not meaningful.

%%%%%=====================
%\subsection{DiPPR objective function}
%\label{sec:DiPPR}
%%%%%=====================

%%%%%=====================
\subsection{Diversity-promoting policy gradients}
\label{sec:opt}
%%%%%=====================

We will use $\theta \in \mathbb{R}^d$ to denote the parameters of a neural network that takes as input $s\in\mathcal{S}$ and outputs a probability distribution over $\mathcal{A}$ with the policy output mapping $\mathcal{Z}:\mathcal{S}\to\mathbb{R}^n$.
For model parameters $\theta$, the action selection distribution as a particular state, $s \in \mathcal{S}$, is denoted by $\pi_\theta(s)$.

The standard gradient loss estimate for policy gradient is given by
\start
\label{eq:gradPG}
    \nabla\mathcal{J}_{\text{PG}}(\pi_{\theta}(s))
    = r(s,a)\nabla\pi_{\theta}(a|s).
\finish
% \andrew{should this actually be:
% \start
% \begin{aligned}
% % \label{eq:gradPG}
%     % PG
%     \nabla_\theta\mathcal{J}_{\textsf{PG}}(\pi_{\theta})
%     &=\sum_{a\in\mathcal{A}}r(s,a)\pi_{\theta}(a|s)\nabla\ln\pi_{\theta}(a|s)\\
%     &=\mathbb{E}_{a\sim\pi_{\theta}(s)}(r(s,a)\nabla\ln\pi_{\theta}(a|s)).
% \end{aligned}
% \finish
% Then when we actually implement this, we estimate the gradient using just the selected action because we don't have $r(s,a)$ for the other actions?
% This is in contrast to the regularizers that don't rely on knowing the rewards.
% In particular, the Policy Gradient Theorem allows us to write (where $p(s)$ is density for $\mathcal{S}$)
% \s
%     \nabla\mathcal{J}(\pi_{\theta})
%     =\int_{\mathcal{S}}\sum_{a\in\mathcal{A}}r(s,a)\nabla\pi_{\theta}(a|s)p(s)\;ds,
% \f
% which we estimate using our samples (say at states $s_1,...,s_T$ with actions $a_1,...,a_T$) as
% \s
%     \nabla\mathcal{J}(\pi_{\theta})
%     \approx\frac{1}{T}\sum_{i=1}^{T}r(s_i,a_i)\nabla\pi_{\theta}(a_i|s_i).
% \f
% Hence, for a single state, we only update the policy based on the selected action.
% (If this is right and we want to include it, I think we can move it to the Policy gradient methods section.)
% }
The gradients of each of the $\varphi$-divergences can be found in Table~\ref{fig:phi-divergences-finite-actions-space}.
Lastly, the gradient of MMD in the contextual bandit setting is
\start 
\begin{aligned}
\label{eq:MMD_gradient}
&\nabla_\theta\textsf{MMD}_{k}^2(\pi_{\theta}(s),u)\\
&\qquad=2\mathbb{E}_{\pi_{\theta}\otimes\pi_{\theta}\otimes u}\Big(\big(k(a,a')-k(a,a^{\star})\big)\nabla_{\theta}\ln\pi_{\theta}(a|s)\Big)\\
%&=\sum_{a}\left(\sum_{a',a^{\star}}(k(a,a')-k(a,a^{\star}))u(a^{\star})\pi_{\theta}(s,a')\right)\nabla_{\theta}\pi_{\theta}(s,a)\\
&\qquad=\sum_{a\in\mathcal{A}}c_{\theta,s,a}(a',a^{\star})\nabla_{\theta}\pi_{\theta}(a|s),
\end{aligned}
\finish
where
\s
c_{\theta,s,a}(a',a^{\star})=\sum_{a',a^{\star}}\big(k(a,a')-k(a,a^{\star})\big)u(a^{\star})\pi_{\theta}(a'|s).
\f
The gradient update from $\nabla_\theta\mathcal{J}_{\textsf{PG}}$ only depends on the gradient based on the action that was selected.
Three of the $\varphi$-divergences, KL, Jensen-Shannon, and Hellinger, as well as MMD have gradients that are weighted sums of the gradients over all of the actions, not just the selected action.
On the other hand, for Total-Variation the gradient only depends on the action whose likelihood is furthest away from the policy $u$, given by~\eqref{eq:uniform_policy}.

When $\pi_{\theta}$ is found using the softmax function, we can further expand all of the above gradients by
\s
\nabla\pi_{\theta}(a|s)
=\pi_{\theta}(a|s)\big[\mathbbm{1}(a=a') - \pi(a'|s)\big]_{a'=1}^n \times \nabla\mathcal{Z}(s).
\f
Using the gradient information given by \eqref{eq:gradPG}, optimal solutions to the $\varphi$-divergence-based diversity-promoting objective, given by (\ref{eq:KLopt}), as well optimal solutions to the MMD-based diversity-promoting objective, given by \eqref{eq:MMDopt}, can then be solved with standard gradient-based methods.

%%%%%=====================
\section{Numerical examples}
\label{sec:examples}
%%%%%=====================
In this section we conduct numerical experiments comparing performance of the RL agents with policy regularization methods described in Section~\ref{sec:dper}.
Specifically, we consider the following agents:
\begin{enumerate}
    \item \texttt{pg}: the default policy gradient agent without any regularization to act as a baseline;
    \item \texttt{pg\_ent}: pg-agent with entropy regularization;
    \item \texttt{pg\_mmd}: pg-agent with MMD regularization;
    \item \texttt{pg\_js}: pg-agent with Jensen-Shannon regularization;
    \item \texttt{pg\_hl}: pg-agent with Hellinger regularization; and
    \item \texttt{pg\_tv}: pg-agent with total variation regularization.
\end{enumerate}
%\andrew{
We chose these algorithms so that we could easily identify the impact that the regularizers have in the absence of additional constraints imposed by other algorithms such as TRPO~\cite{schulman2015trust} or PPO~\cite{schulman2017proximal}.
%}
%\anton{Do we still need these?}
%\andrew{I don't think we need them, but maybe they're good to have because their so popular?}

The agents and regularizer losses are manually implemented in TensorFlow and the network training is performed with Adam optimizer with the default hyperparameters.
For all of our algorithm configurations, we use a batch size of 100.
An agent policy is parameterized by a 2-layer feed-forward neural network with 32 nodes on each layer.
For each regularized agent we perform a hyperparameter search to determine the appropriate value of the regularization coefficient.

We deploy the agents on various personalization tasks that are given by contextual bandit environments.
For each agent and environment we report the following metrics, computed over the test set: the agent reward, the policy entropy, and the action selection histogram.
For the simplicity of presentation, the histograms are sorted to emphasize the agent's action distribution over the test set.

The presented examples are performed using Python3.8 with Tensorflow~2.12 on personal laptops.
The source code reproducing the given experiments is available at~\url{https://github.com/acstarnes/wain23-policy-regularization}.
%The source code reproducing the given experiments will be made publicly available at the time of publication.

%%%%%=====================
\subsection{MNIST Environment}
\begin{figure*}[t]
    \centering
    \includegraphics[width=.8\linewidth]{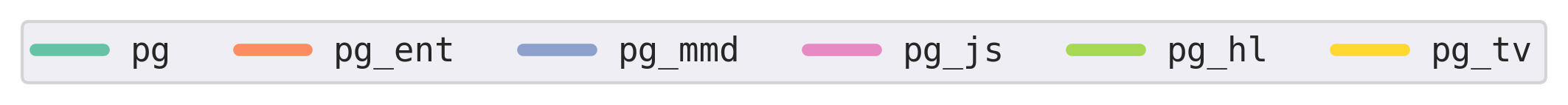}
    \\
    \begin{subfigure}{.49\linewidth}
        \includegraphics[width=\linewidth]{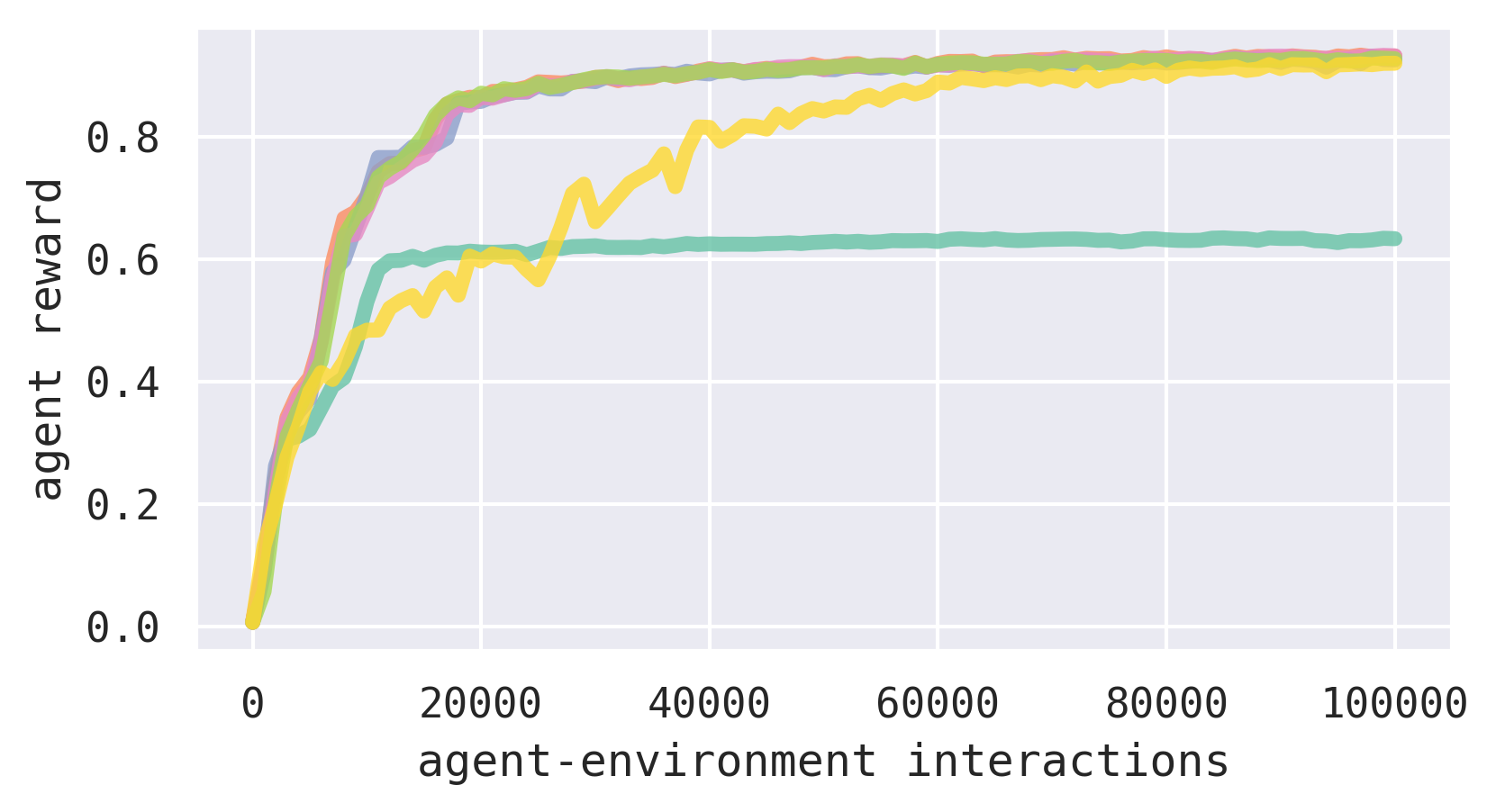}
        \caption{Agent reward}
    \end{subfigure}
    \begin{subfigure}{.49\linewidth}
        \includegraphics[width=\linewidth]{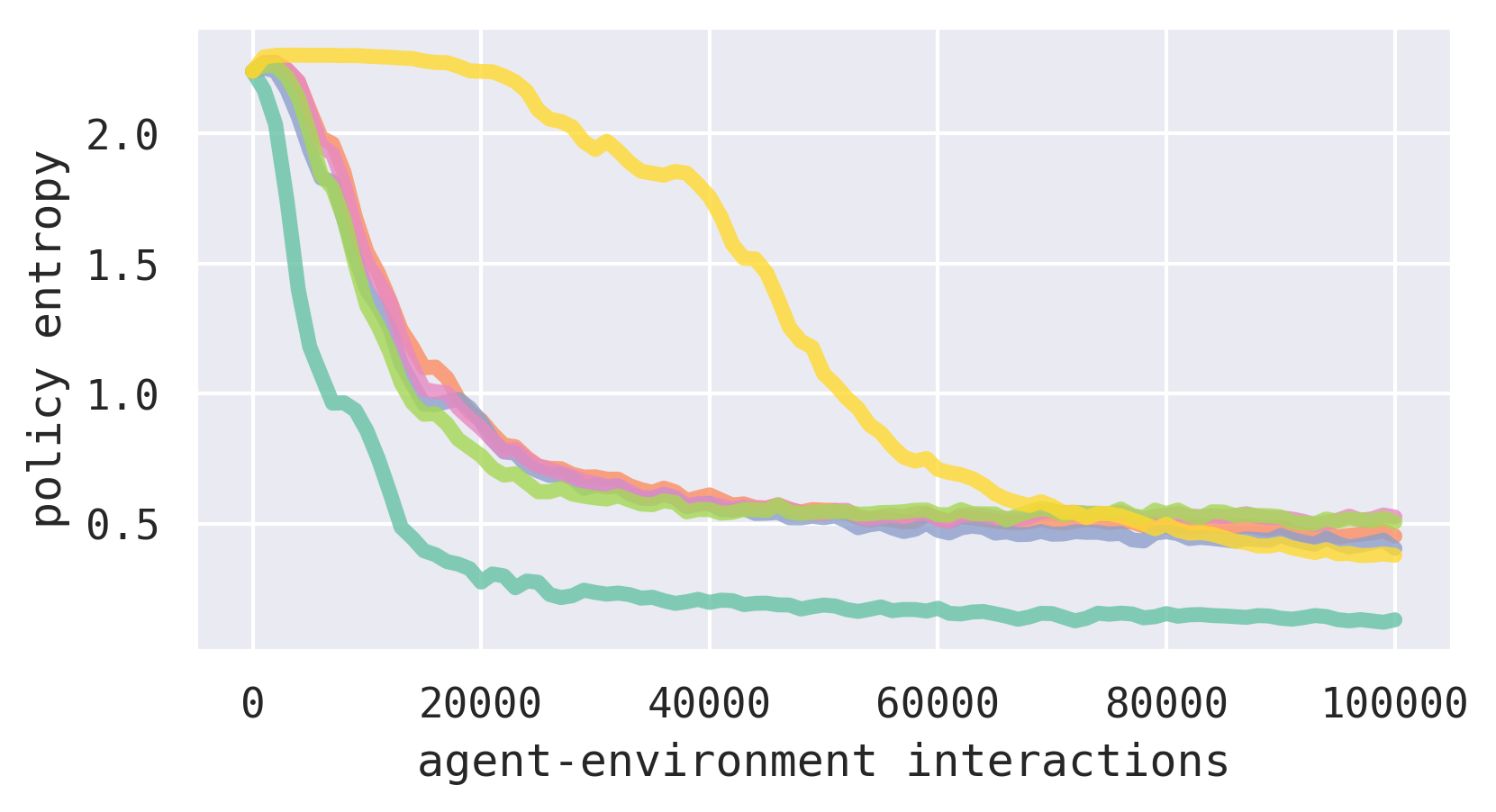}
        \caption{Policy entropy}
    \end{subfigure}
    \\
    \begin{subfigure}{.32\linewidth}
        \includegraphics[width=\linewidth]{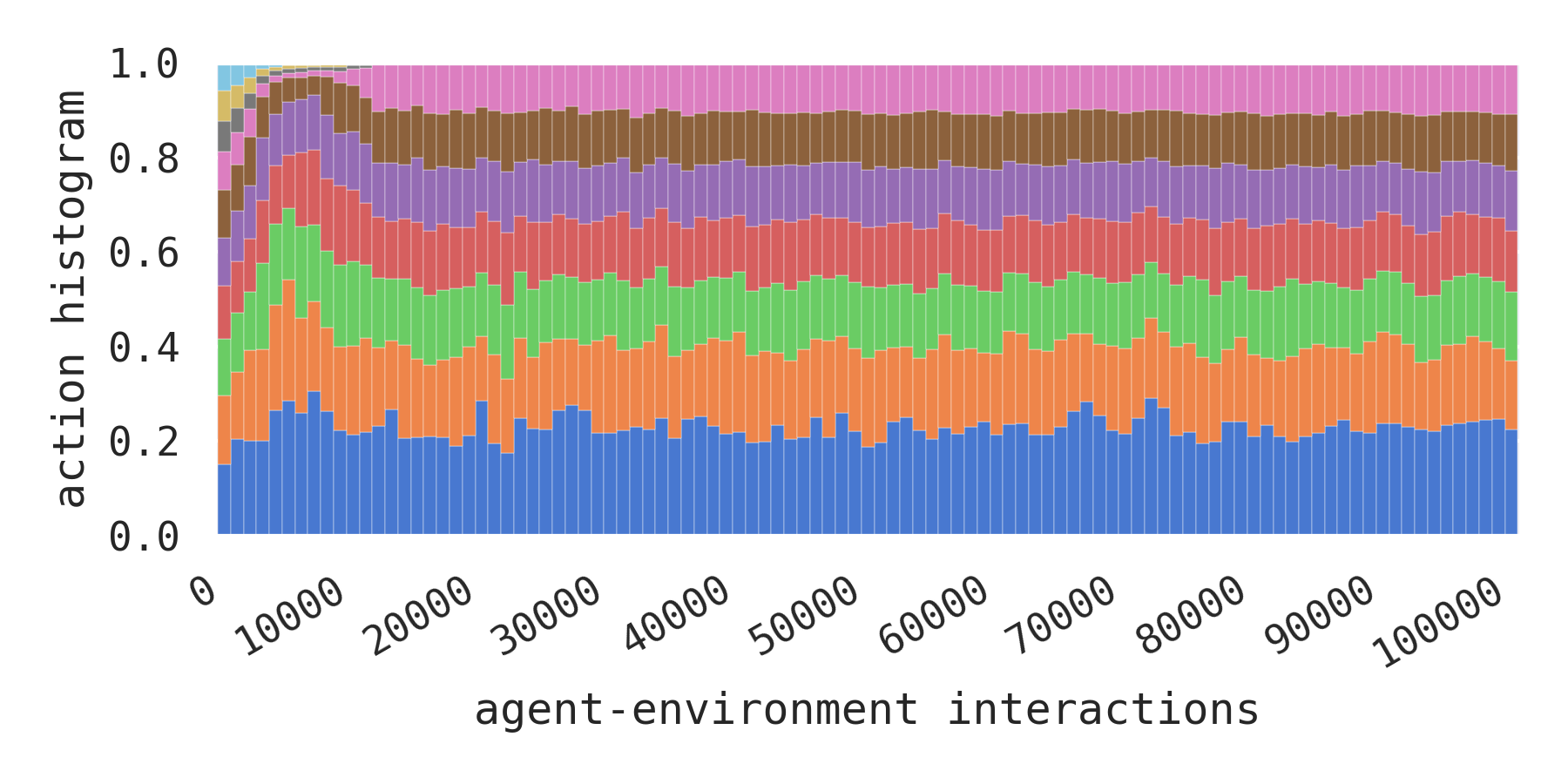}
        \caption{No regularization}
    \end{subfigure}
    \begin{subfigure}{.32\linewidth}
        \includegraphics[width=\linewidth]{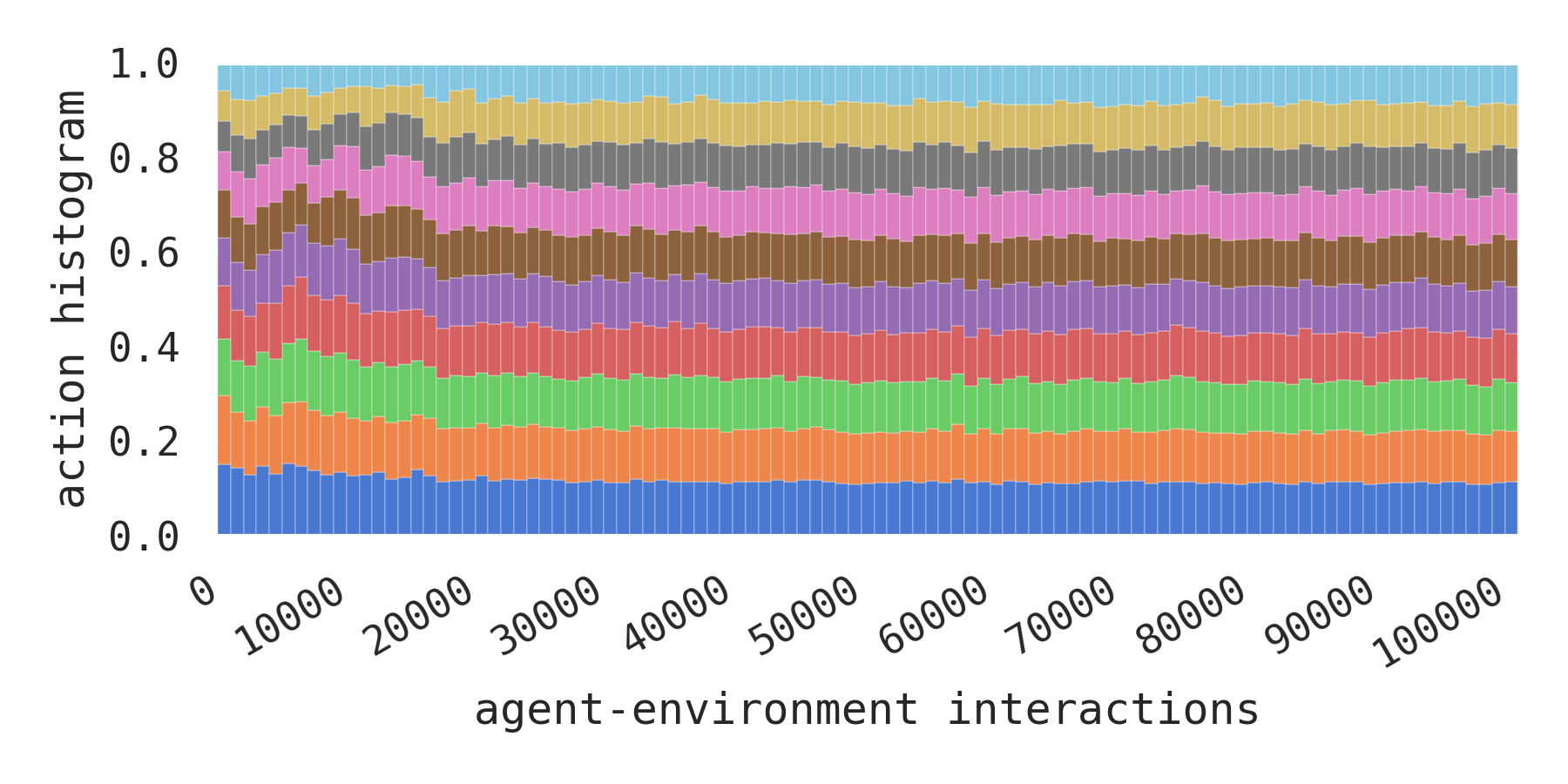}
        \caption{Entropy regularization}
    \end{subfigure}
    \begin{subfigure}{.32\linewidth}
        \includegraphics[width=\linewidth]{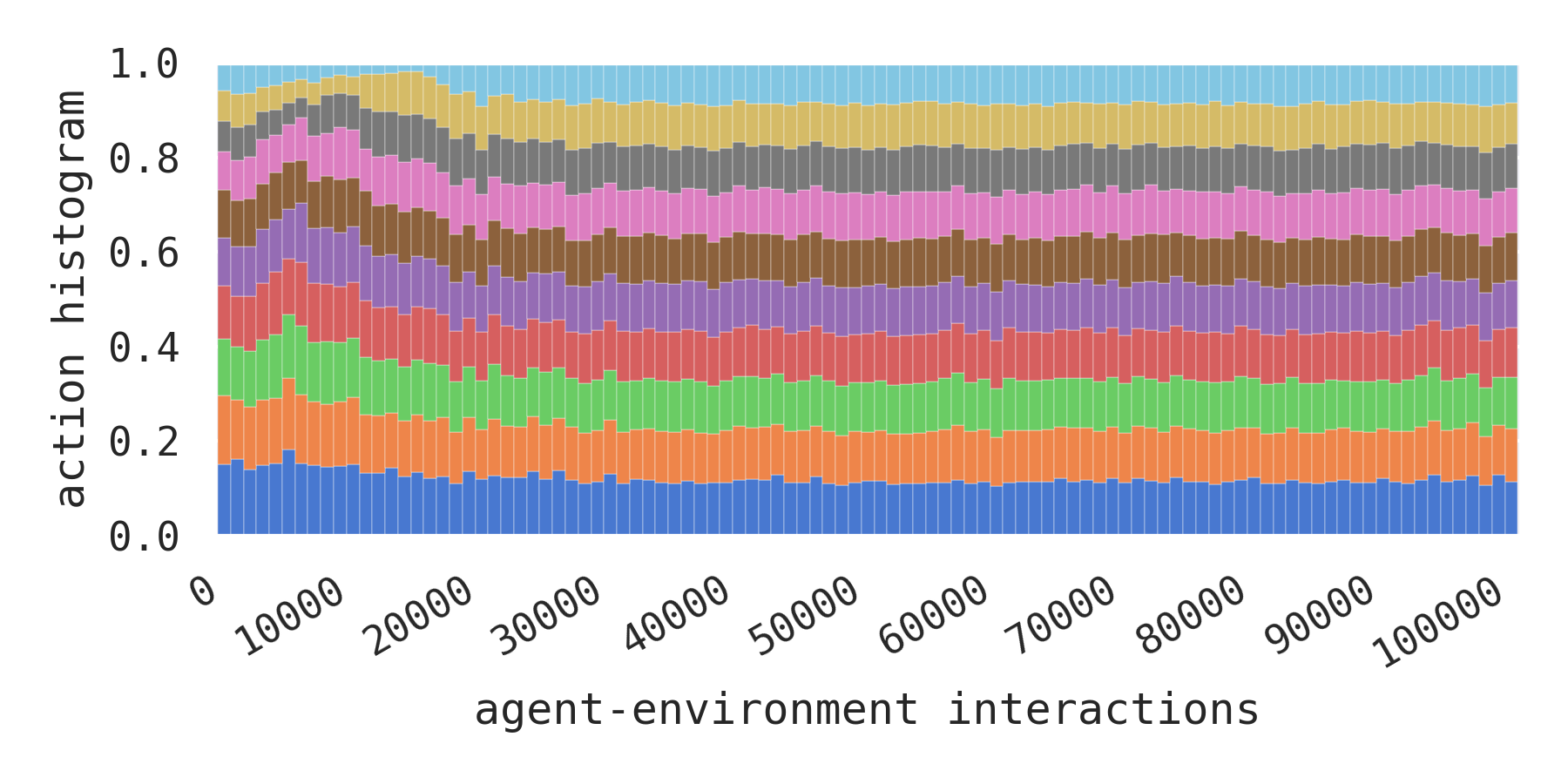}
        \caption{MMD regularization}
    \end{subfigure}
    \\
    \begin{subfigure}{.32\linewidth}
        \includegraphics[width=\linewidth]{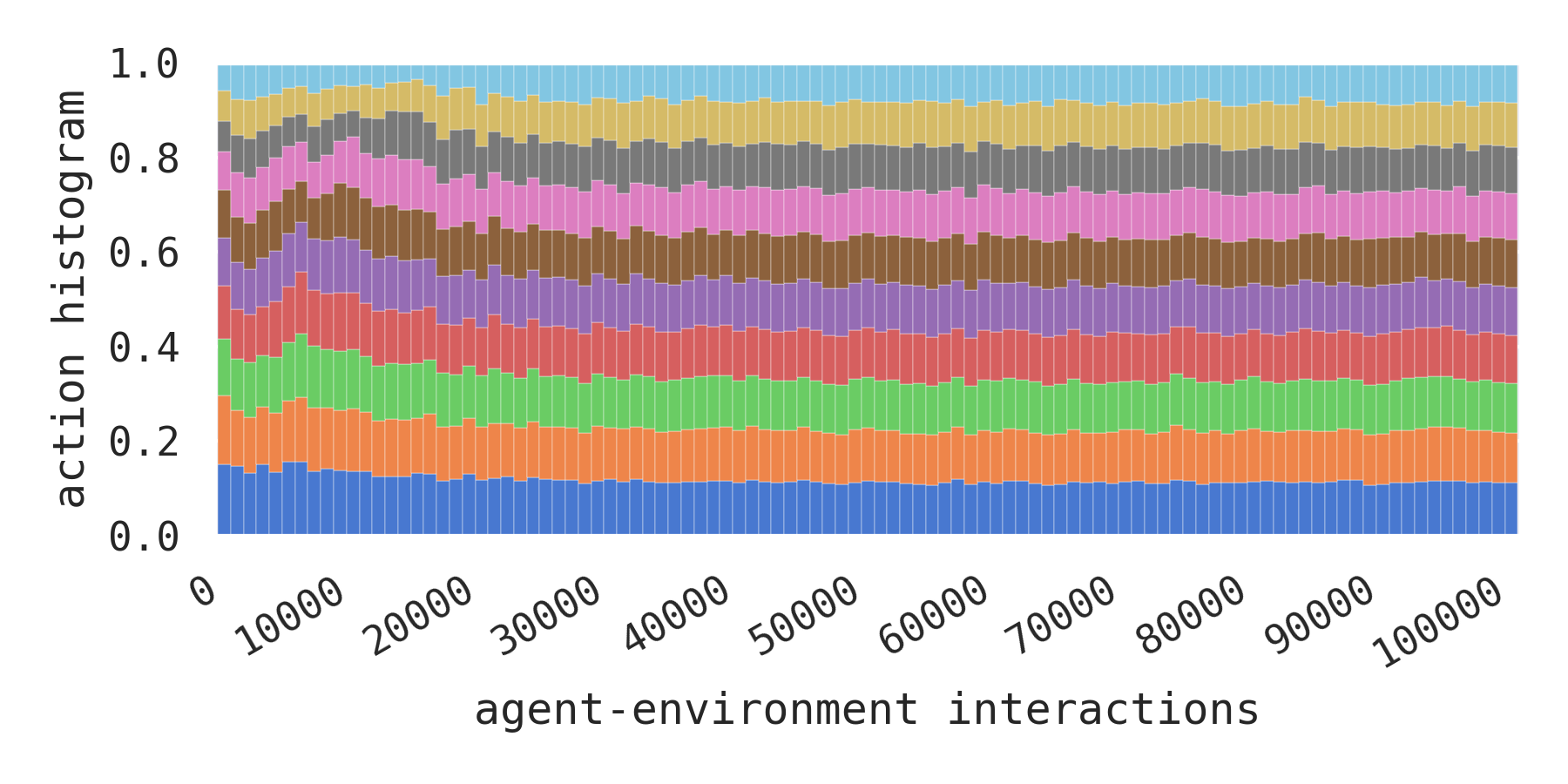}
        \caption{Jensen-Shannon regularization}
    \end{subfigure}
    \begin{subfigure}{.32\linewidth}
        \includegraphics[width=\linewidth]{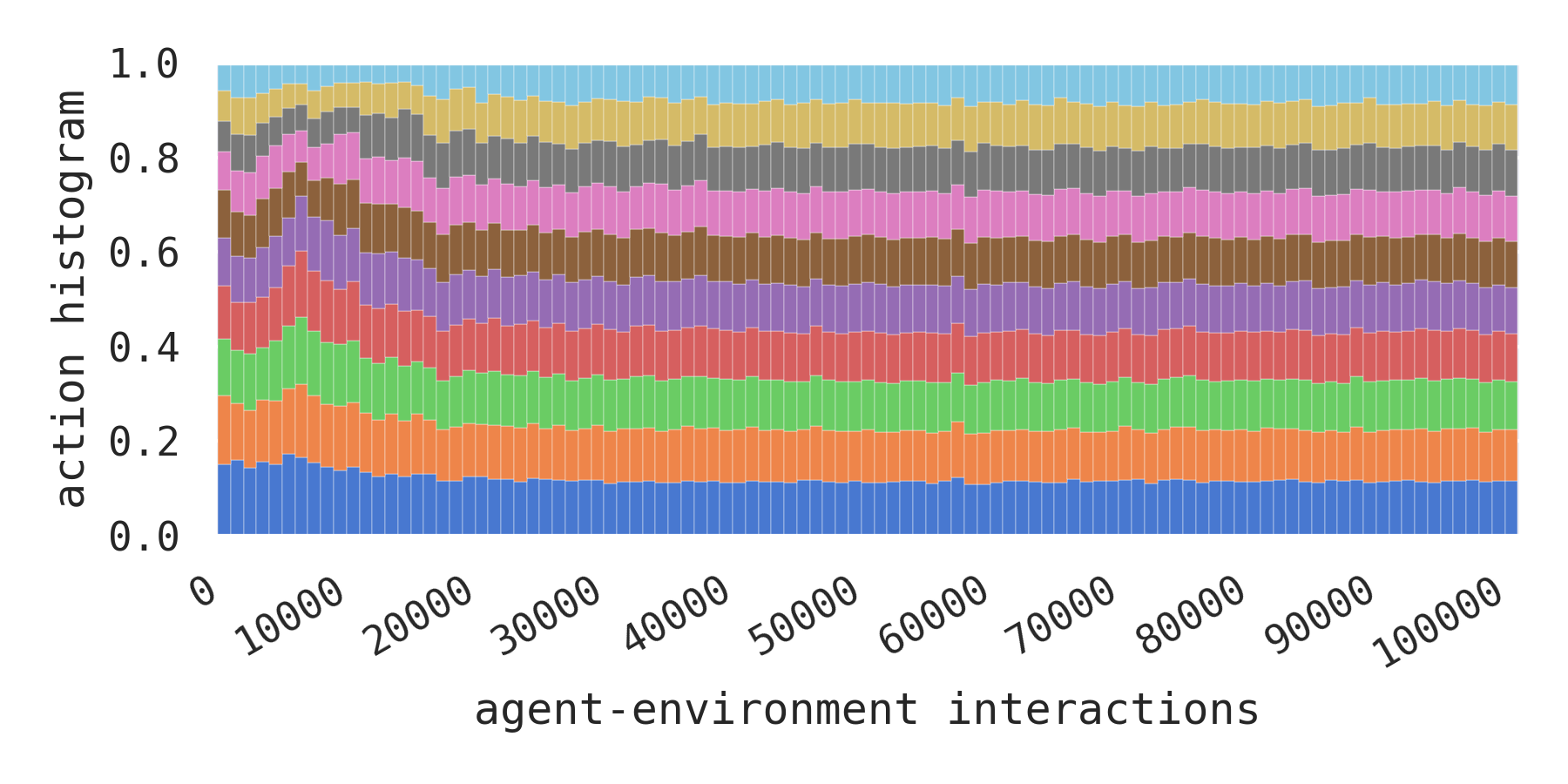}
        \caption{Hellinger regularization}
    \end{subfigure}
    \begin{subfigure}{.32\linewidth}
        \includegraphics[width=\linewidth]{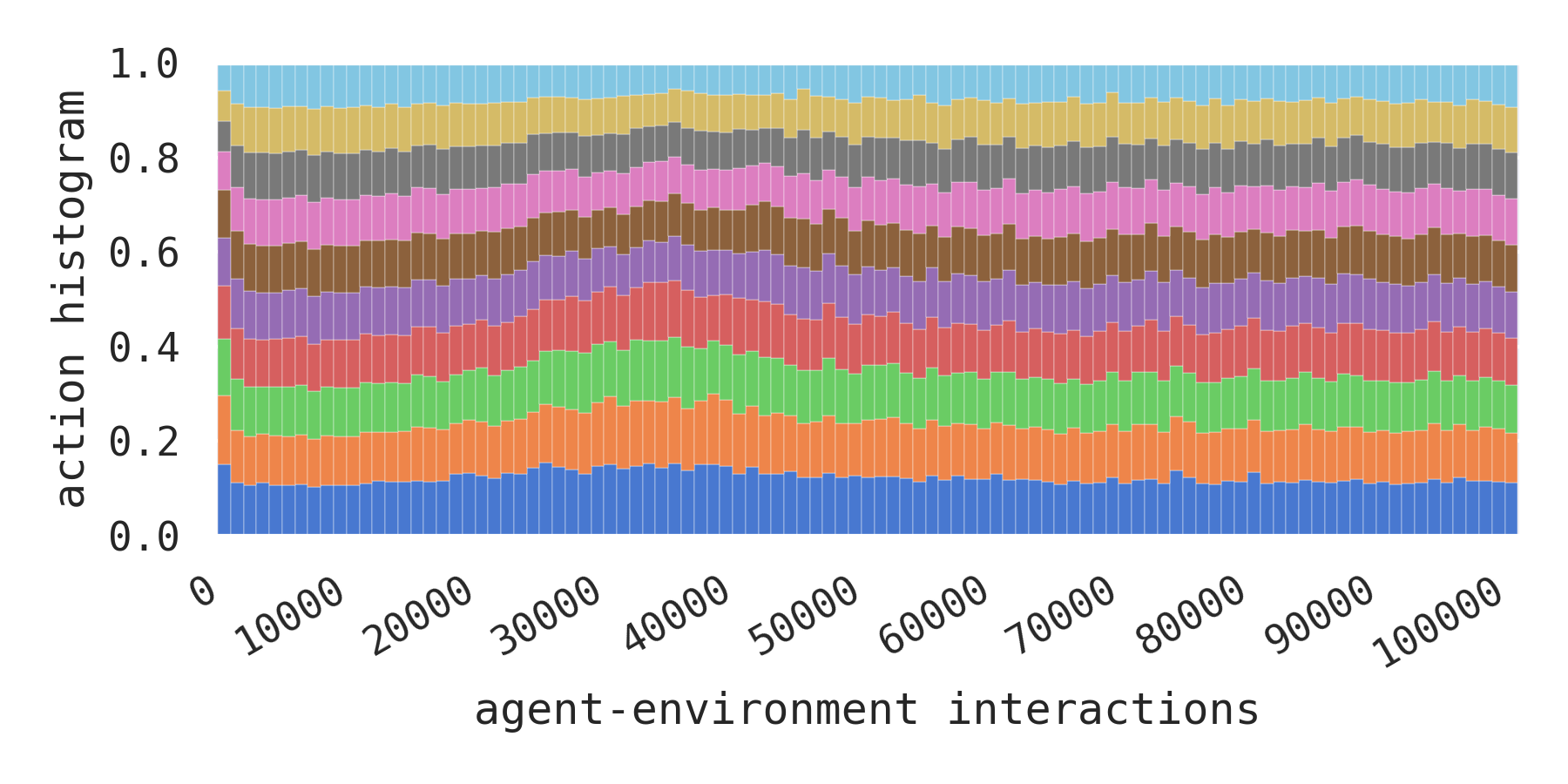}
        \caption{Total variation regularization}
    \end{subfigure}
    \caption{Results of image classification experiment on MNIST environment.}
    \label{fig:mnist}
\end{figure*}

We use MNIST dataset\footnote{\url{http://yann.lecun.com/exdb/mnist/}} to create a contextual bandit environment, as done in \cite{dudik2011doubly, swaminathan2015counterfactual, chen2019surrogate, Dereventsov_2022}.
In this formulation the images act as observations and the labels act as the actions that agents can take.
The reward for selecting the correct label is 1 and $-\nicefrac{1}{9}$ for an incorrect classification.
Defining the reward function this way means that the expected return for the uniformly random policy is $0$ and for the optimal policy is $1$.
The agent reward, policy entropy, and action selection histograms for the various regularizers on MNIST environment are shown in Figure~\ref{fig:mnist}.

We note that all regularized agents solve the environment and demonstrate a comparable performance while outperforming the baseline agent.
From the action selection histogram we observe that the unregularized policy gradient agent only selects $7$ out of $10$ actions, which results in the agent reward value plateauing at about $0.6$.
In contrast, all of the regularized agents maintain diverse action selection throughout the training process and achieve reward values close to $1$, which indicates fully learning the environment.

%%%%%=====================
\subsection{CIFAR10 Environment}
\begin{figure*}[t]
    \centering
    \includegraphics[width=.8\linewidth]{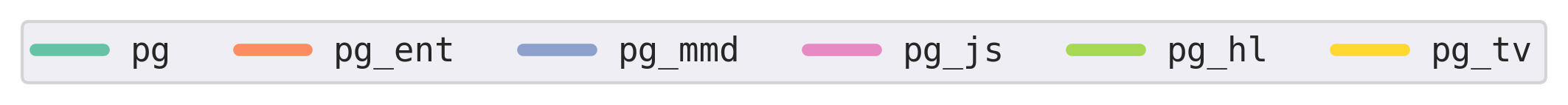}
    \\
    \begin{subfigure}{.49\linewidth}
        \includegraphics[width=\linewidth]{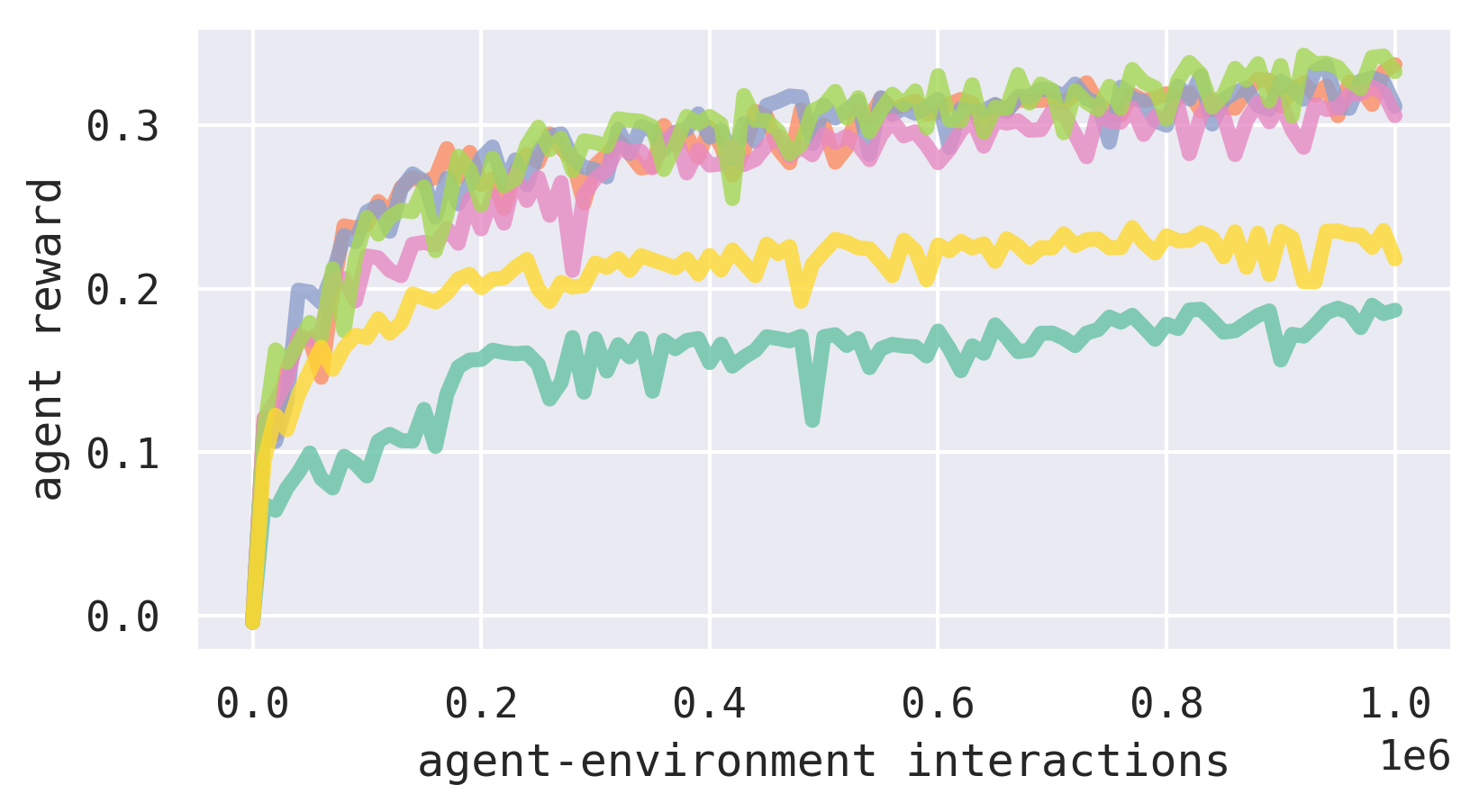}
        \caption{Agent reward}
    \end{subfigure}
    \begin{subfigure}{.49\linewidth}
        \includegraphics[width=\linewidth]{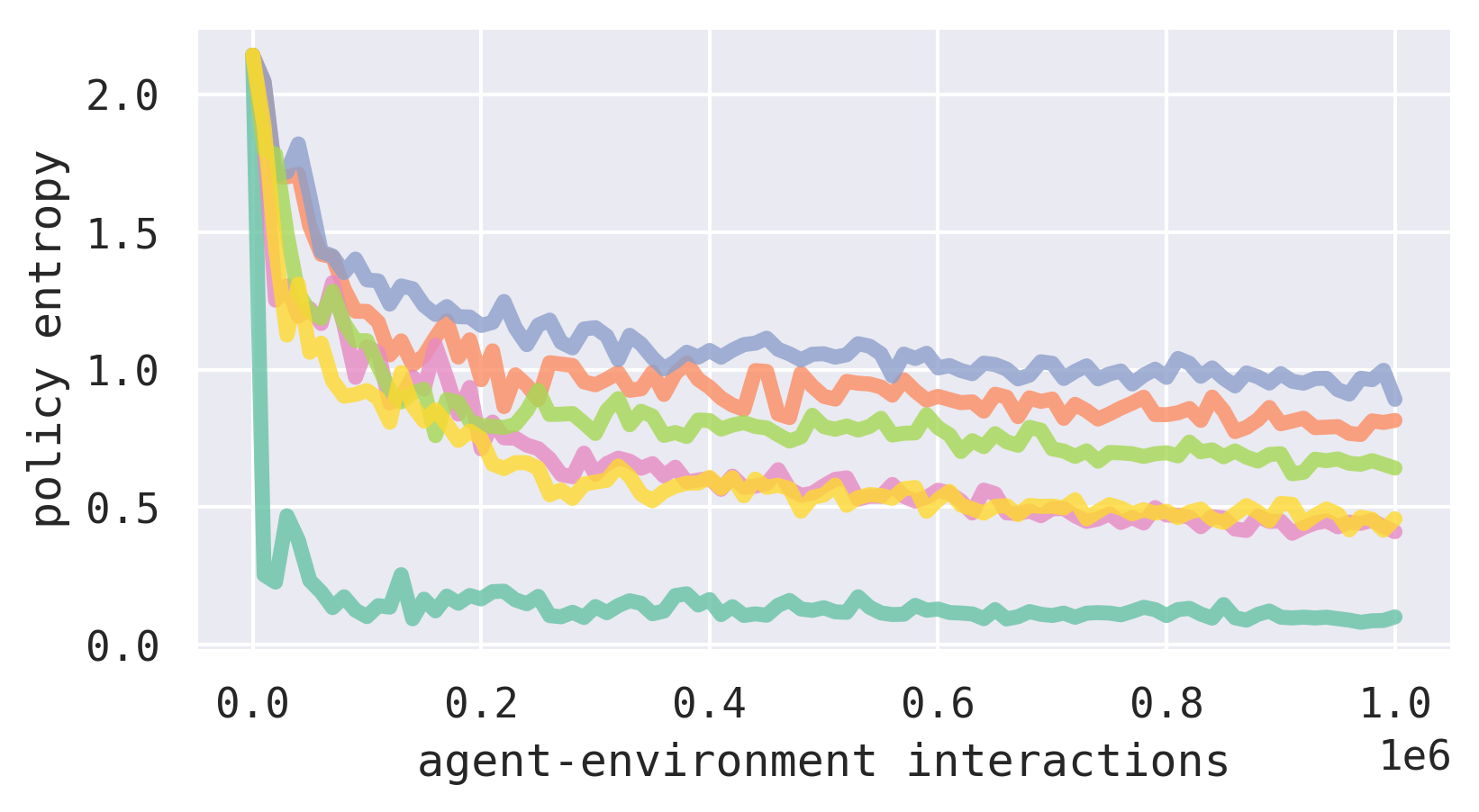}
        \caption{Policy entropy}
    \end{subfigure}
    \\
    \begin{subfigure}{.32\linewidth}
        \includegraphics[width=\linewidth]{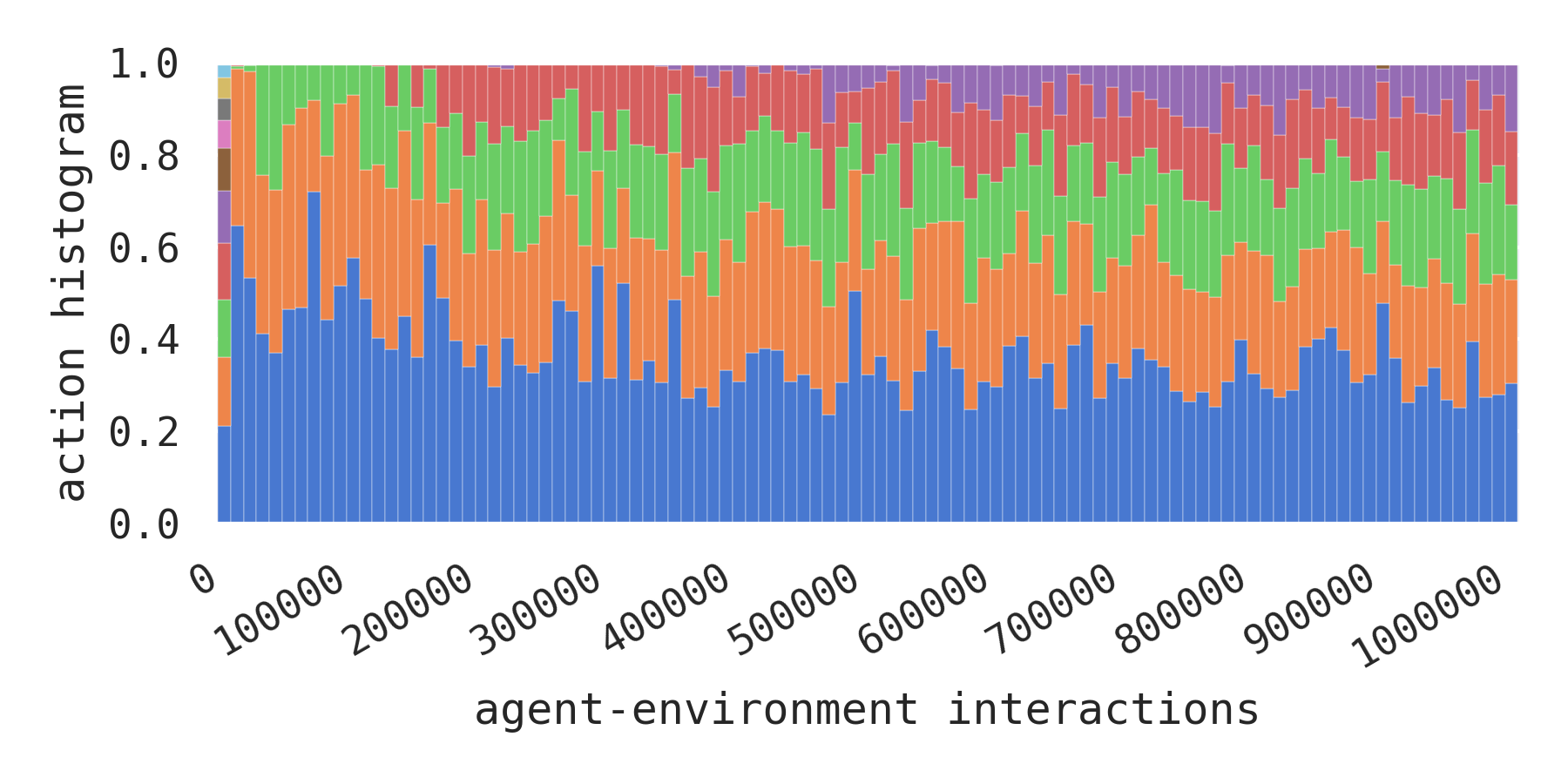}
        \caption{No regularization}
    \end{subfigure}
    \begin{subfigure}{.32\linewidth}
        \includegraphics[width=\linewidth]{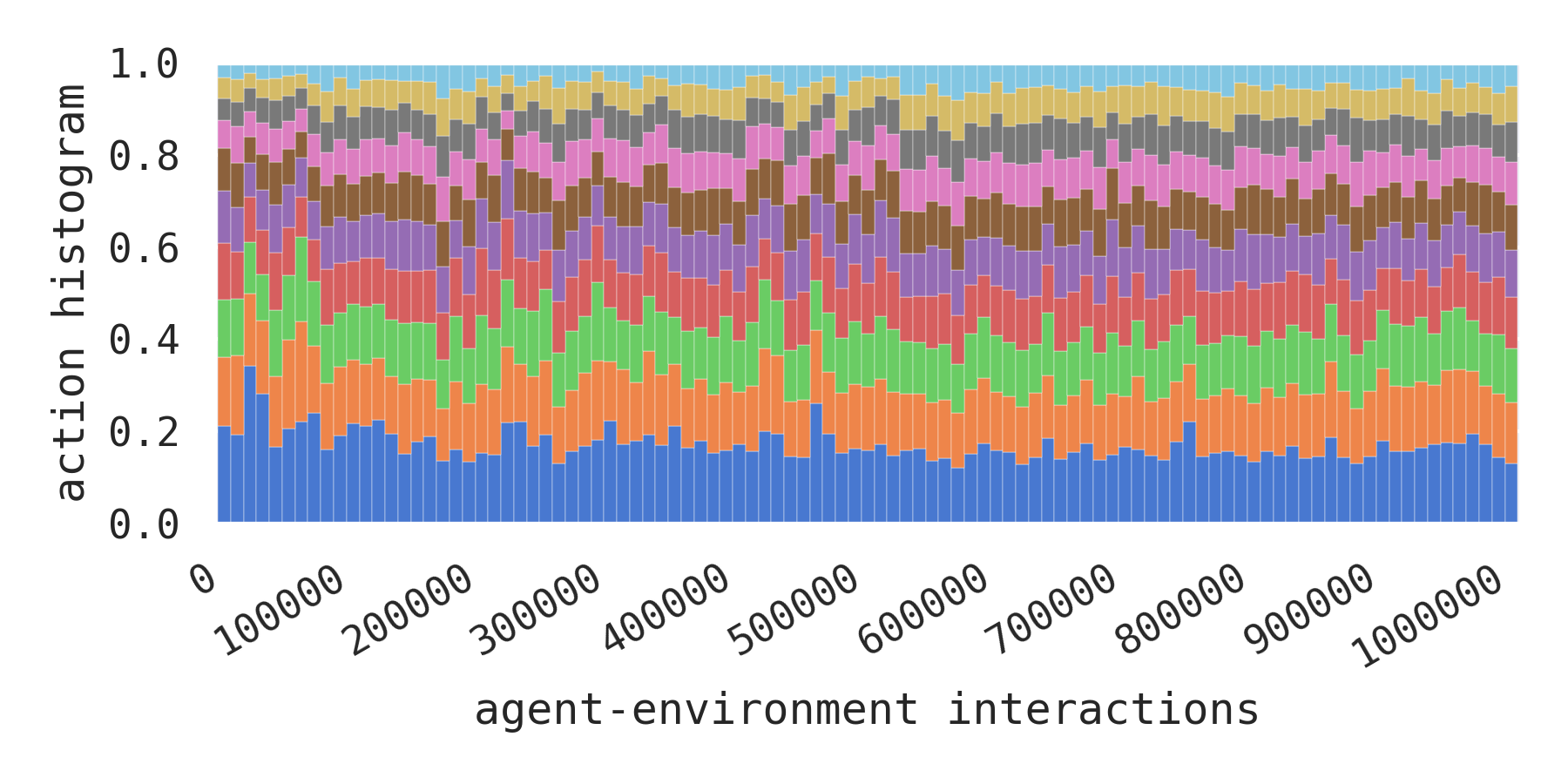}
        \caption{Entropy regularization}
    \end{subfigure}
    \begin{subfigure}{.32\linewidth}
        \includegraphics[width=\linewidth]{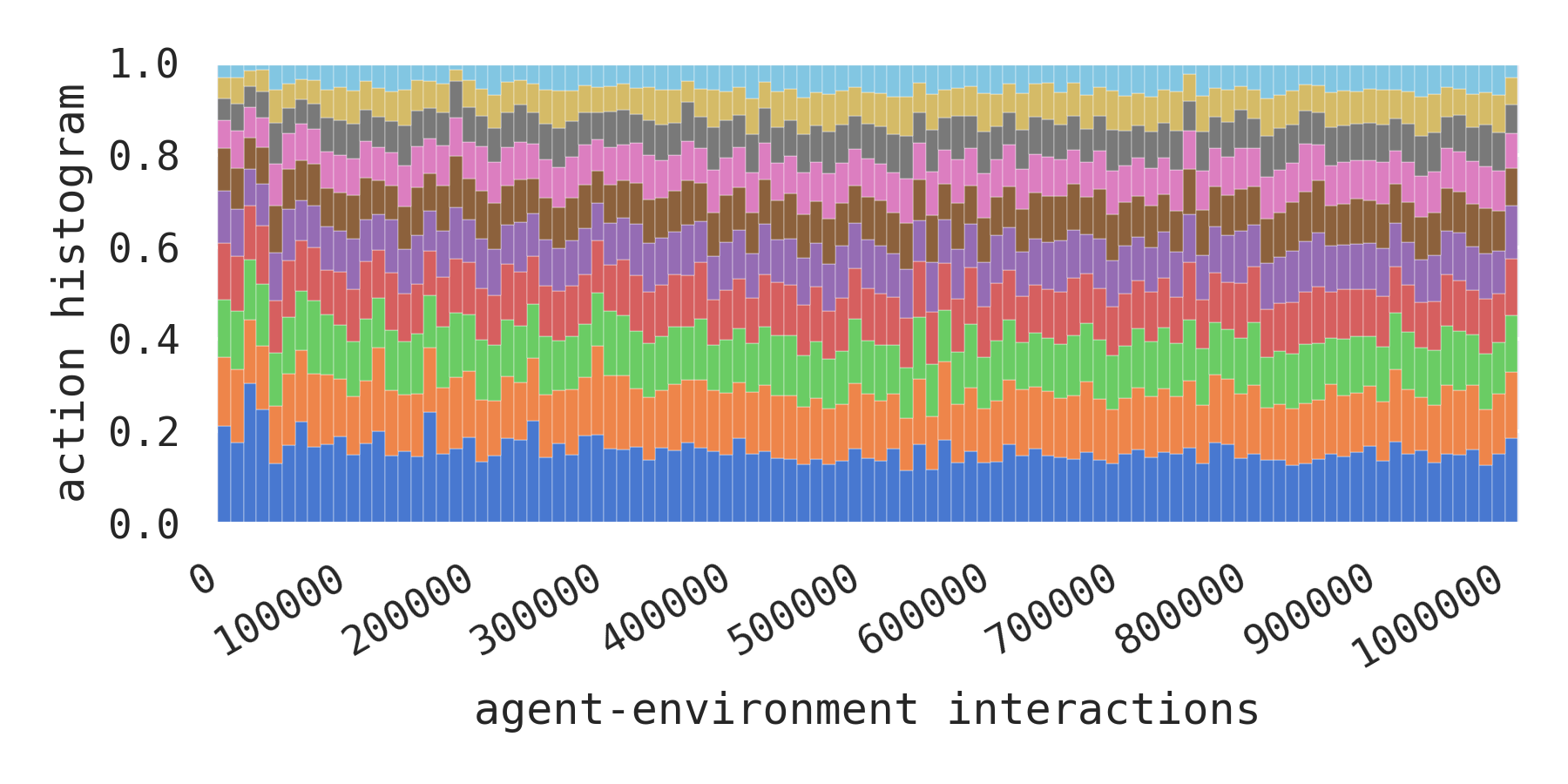}
        \caption{MMD regularization}
    \end{subfigure}
    \\
    \begin{subfigure}{.32\linewidth}
        \includegraphics[width=\linewidth]{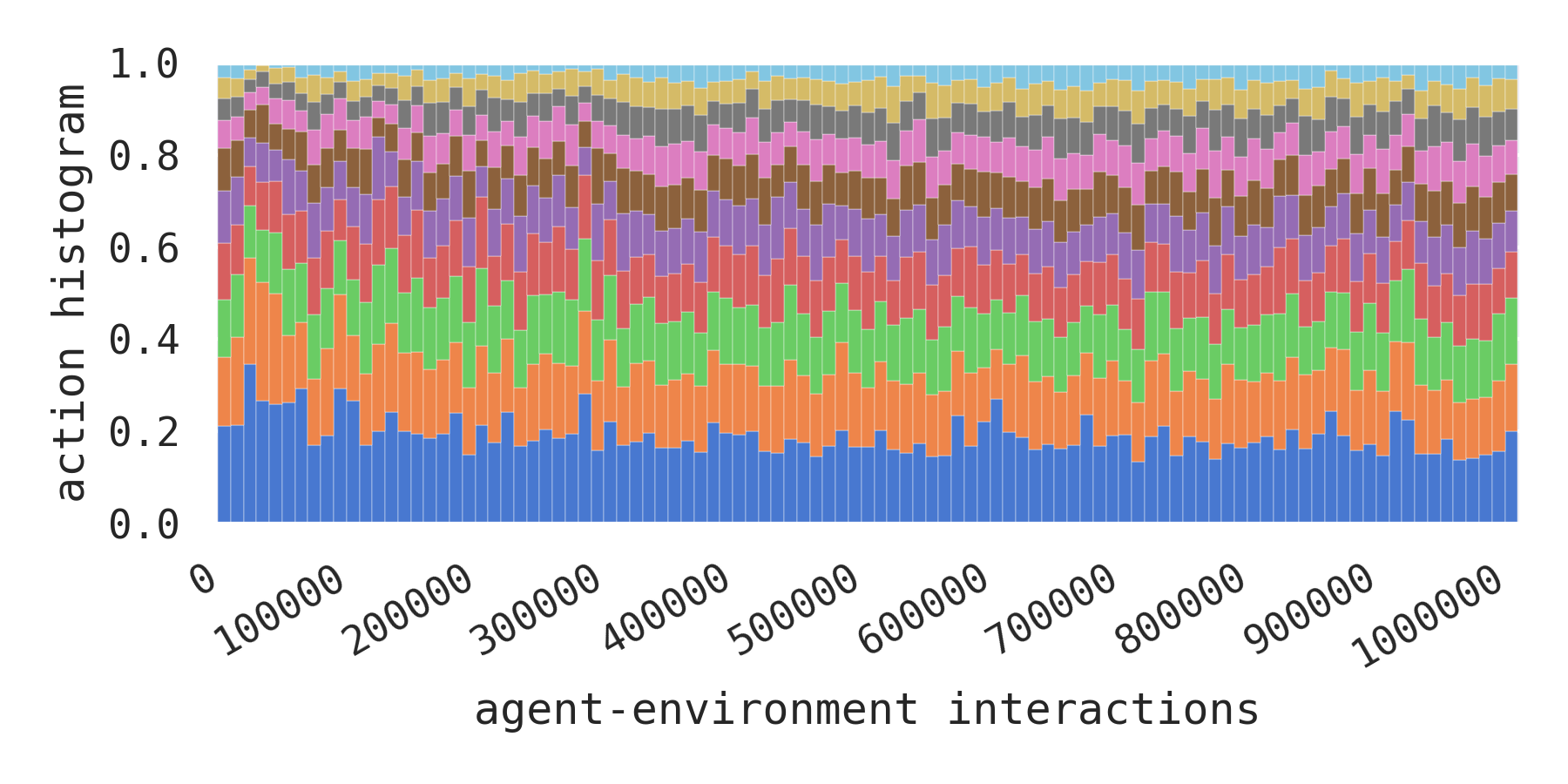}
        \caption{Jensen-Shannon regularization}
    \end{subfigure}
    \begin{subfigure}{.32\linewidth}
        \includegraphics[width=\linewidth]{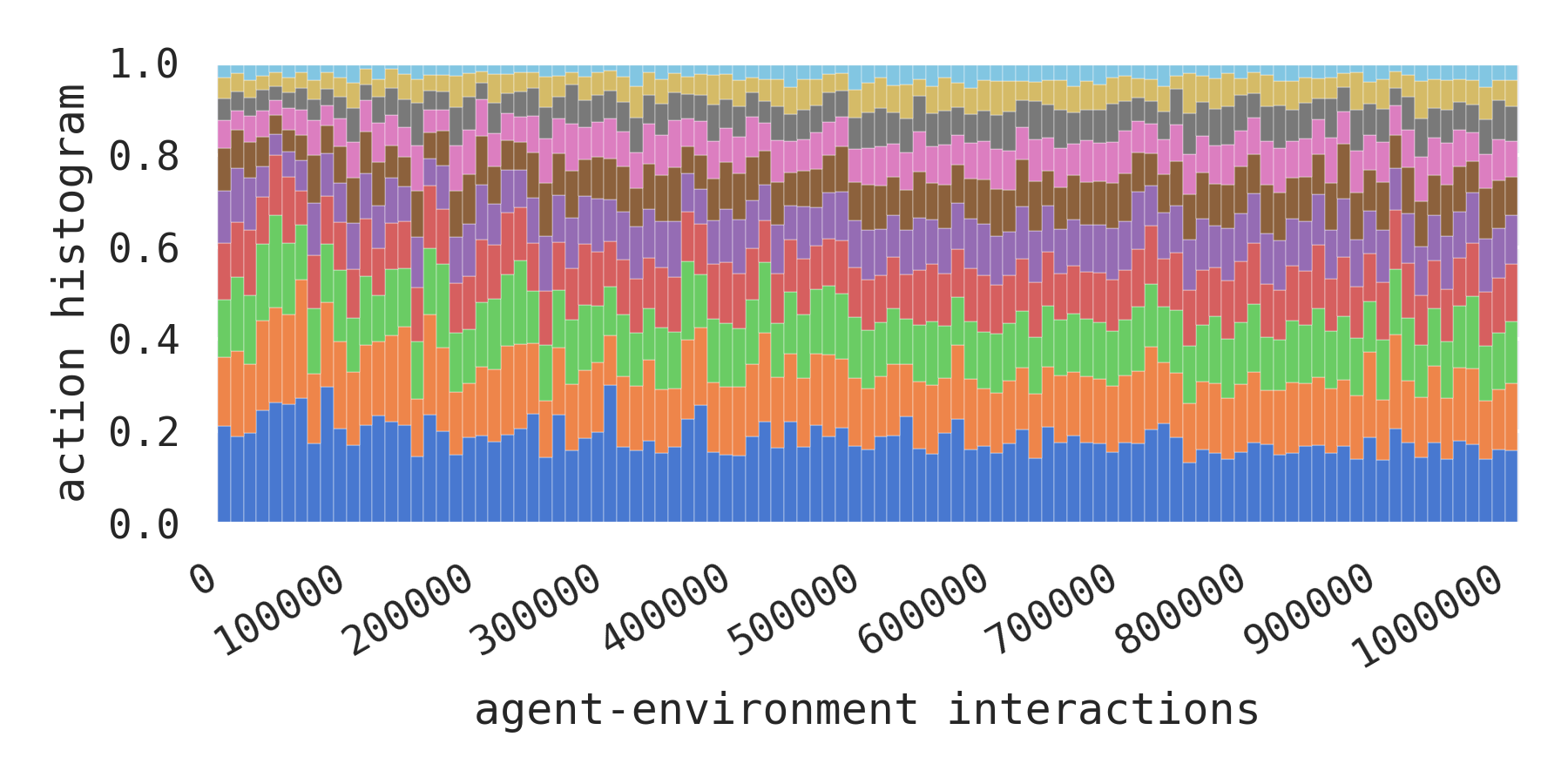}
        \caption{Hellinger regularization}
    \end{subfigure}
    \begin{subfigure}{.32\linewidth}
        \includegraphics[width=\linewidth]{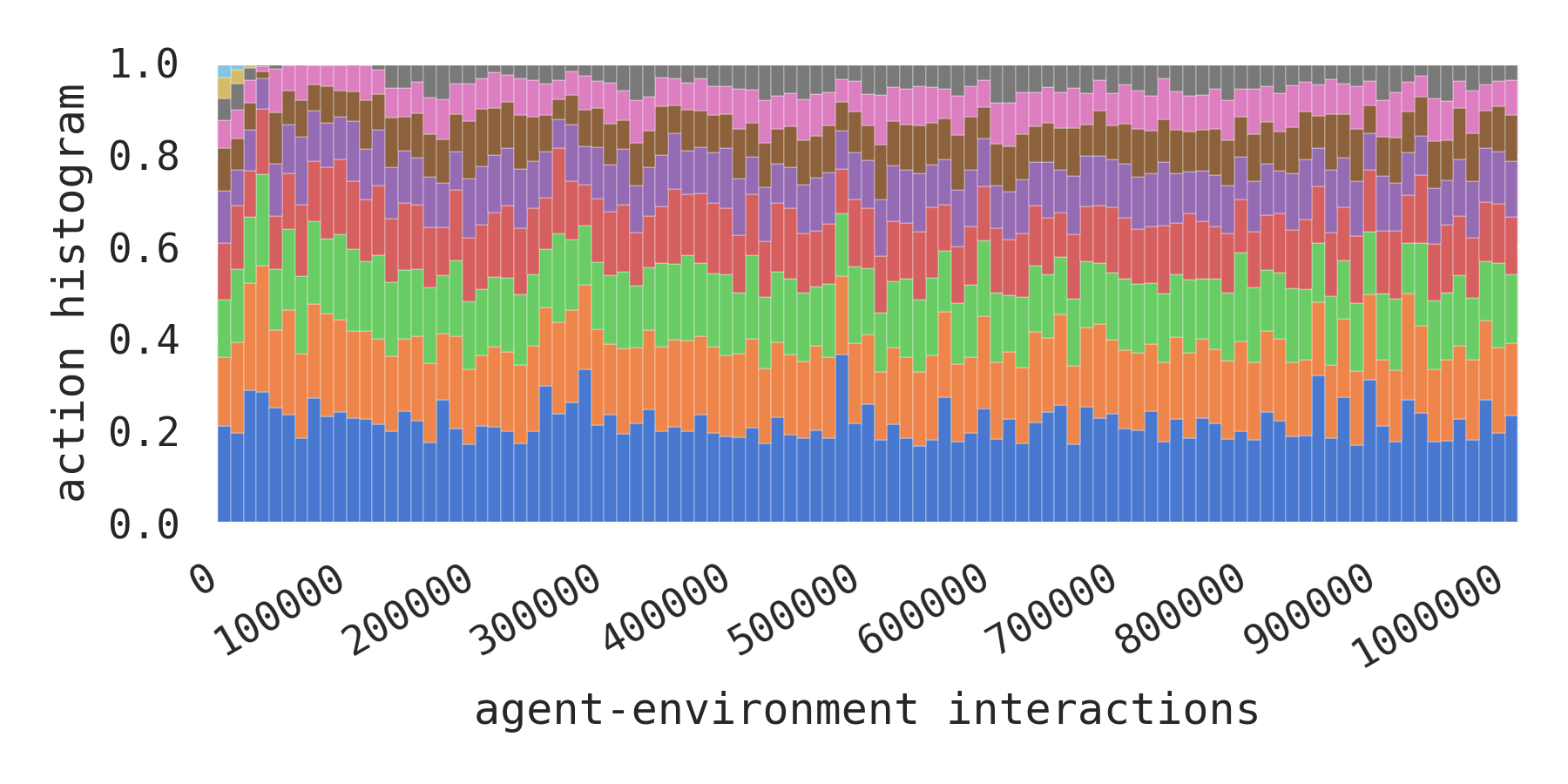}
        \caption{Total variation regularization}
    \end{subfigure}
    \caption{Results of image classification experiment on CIFAR10 environment.}
    \label{fig:cifar10}
\end{figure*}

As in the previous example, we use CIFAR10 dataset\footnote{\url{https://www.cs.toronto.edu/~kriz/cifar.html}} to create a contextual bandit environment.
The agent reward, policy entropy, and action selection histograms for the various regularizers on CIFAR10 environment are shown in Figure~\ref{fig:cifar10}.

Unlike the previous example, the agents fail to fully solve the environment in this case due to the increased complexity of CIFAR10 dataset.
In fact, even the most successful agents achieve reward values of only about $0.3$, which roughly equates to a $35\%$ classification accuracy on CIFAR10 dataset.
Such poor performance is due to the constrained network architecture and the contextual bandit formulation of the problem.
Nonetheless, the advantage of regularized agents is evident, both from policy reward and entropy perspectives.
In particular, we observe that most of the regularized agents are able to maintain diverse (albeit unbalanced) action selection, while the baseline agent only selects $5$ out of $10$ available actions.

%%%%%=====================
\subsection{Spotify Environment}
\begin{figure*}[t]
    \centering
    \includegraphics[width=.8\linewidth]{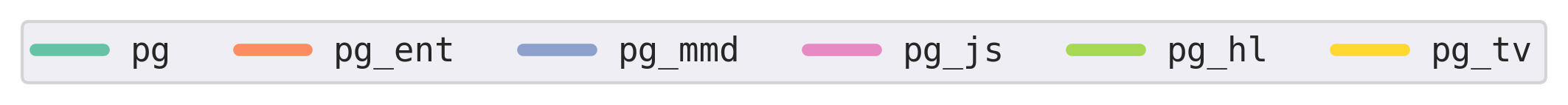}
    \\
    \begin{subfigure}{.49\linewidth}
        \includegraphics[width=\linewidth]{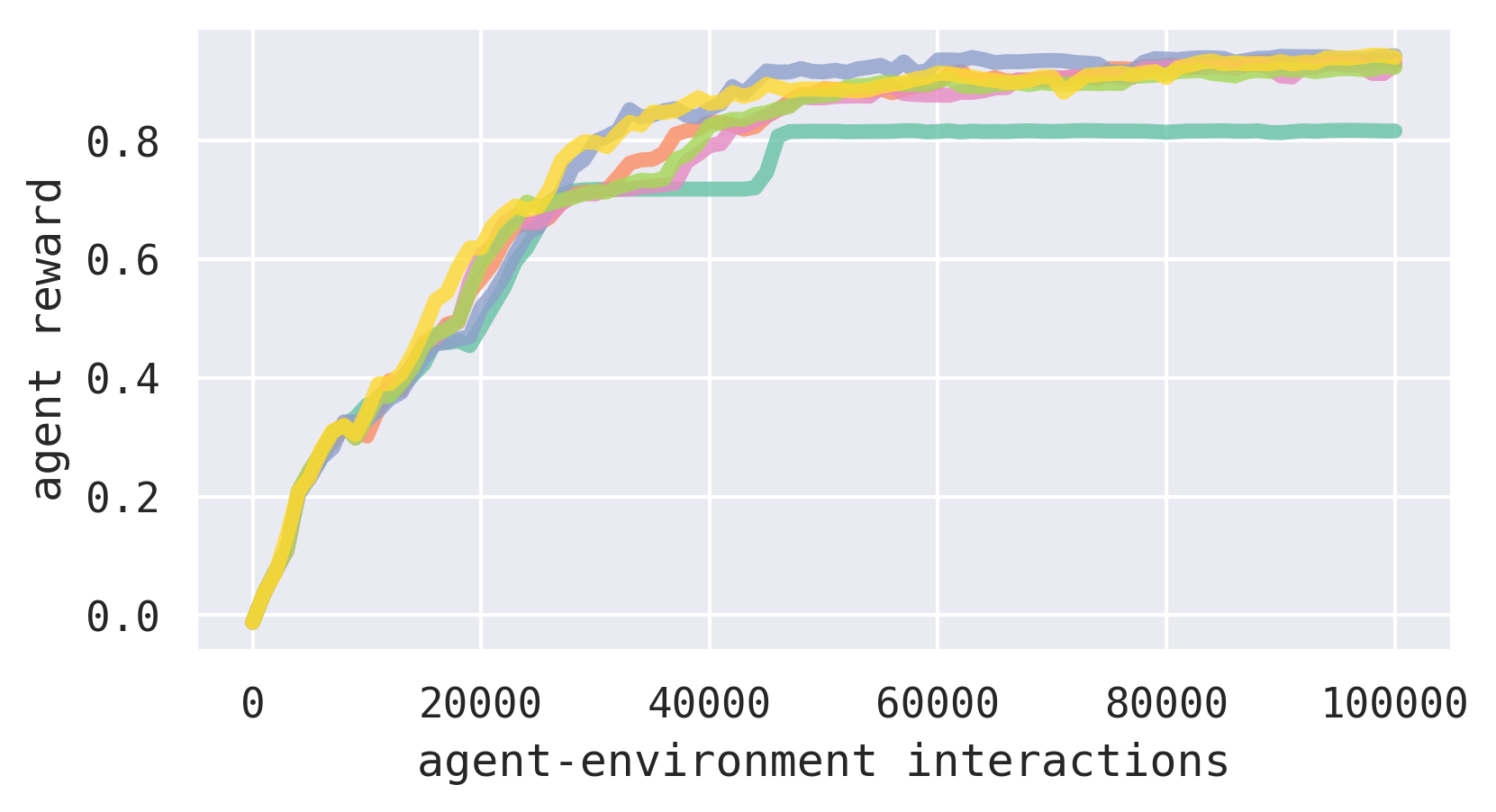}
        \caption{Agent reward}
    \end{subfigure}
    \begin{subfigure}{.49\linewidth}
        \includegraphics[width=\linewidth]{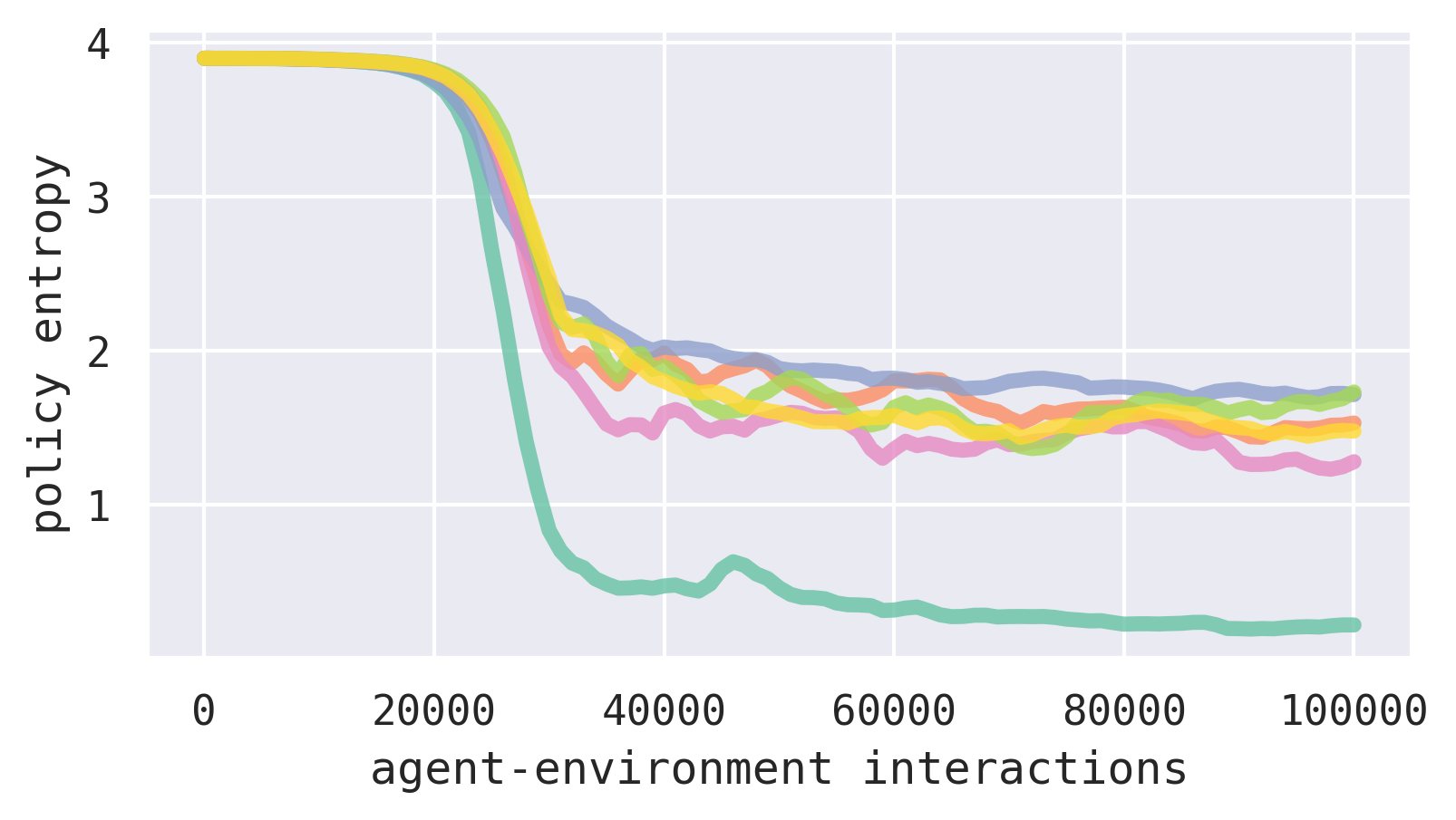}
        \caption{Policy entropy}
    \end{subfigure}
    \\
    \begin{subfigure}{.32\linewidth}
        \includegraphics[width=\linewidth]{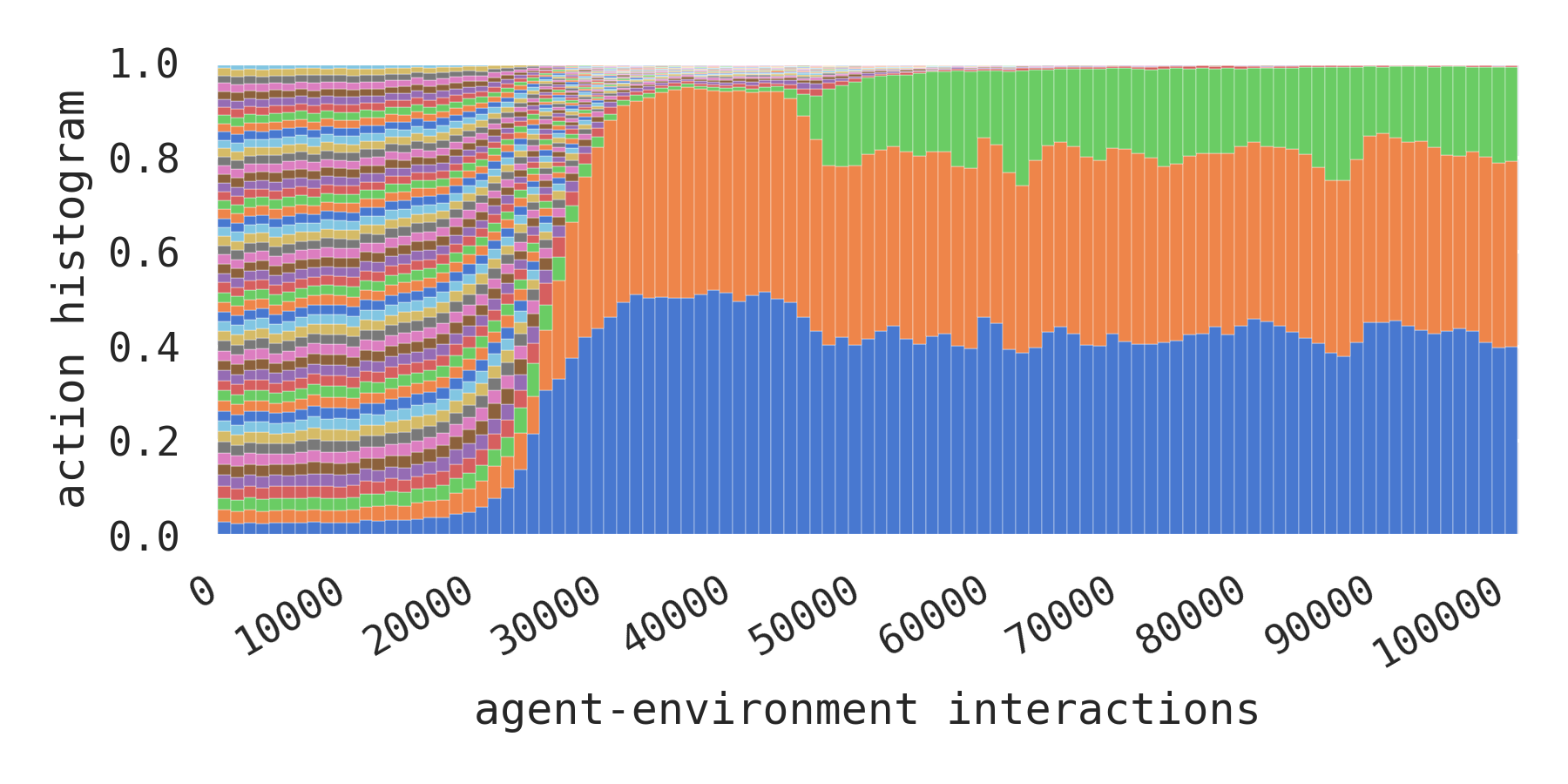}
        \caption{No regularization}
    \end{subfigure}
    \begin{subfigure}{.32\linewidth}
        \includegraphics[width=\linewidth]{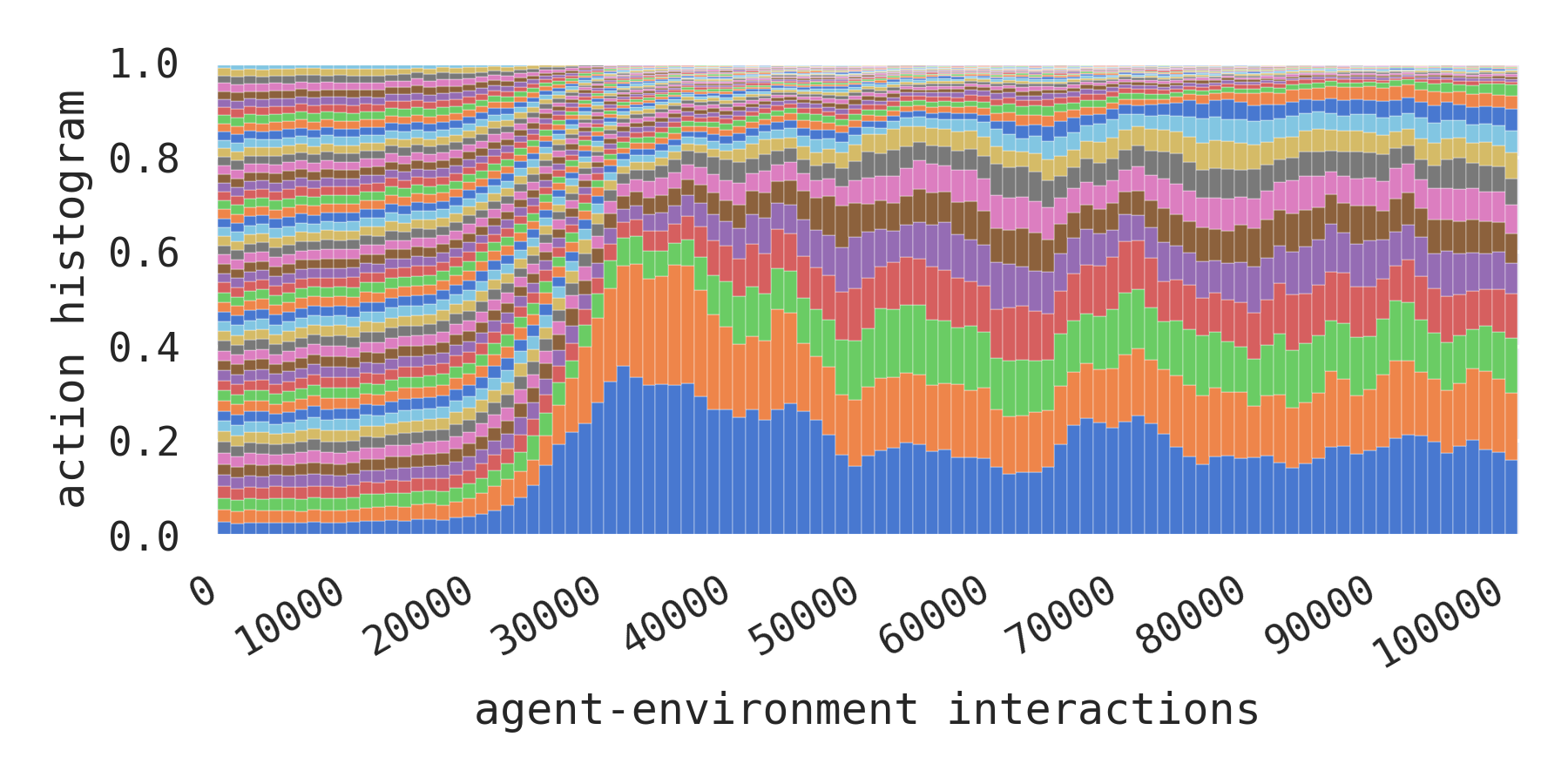}
        \caption{Entropy regularization}
    \end{subfigure}
    \begin{subfigure}{.32\linewidth}
        \includegraphics[width=\linewidth]{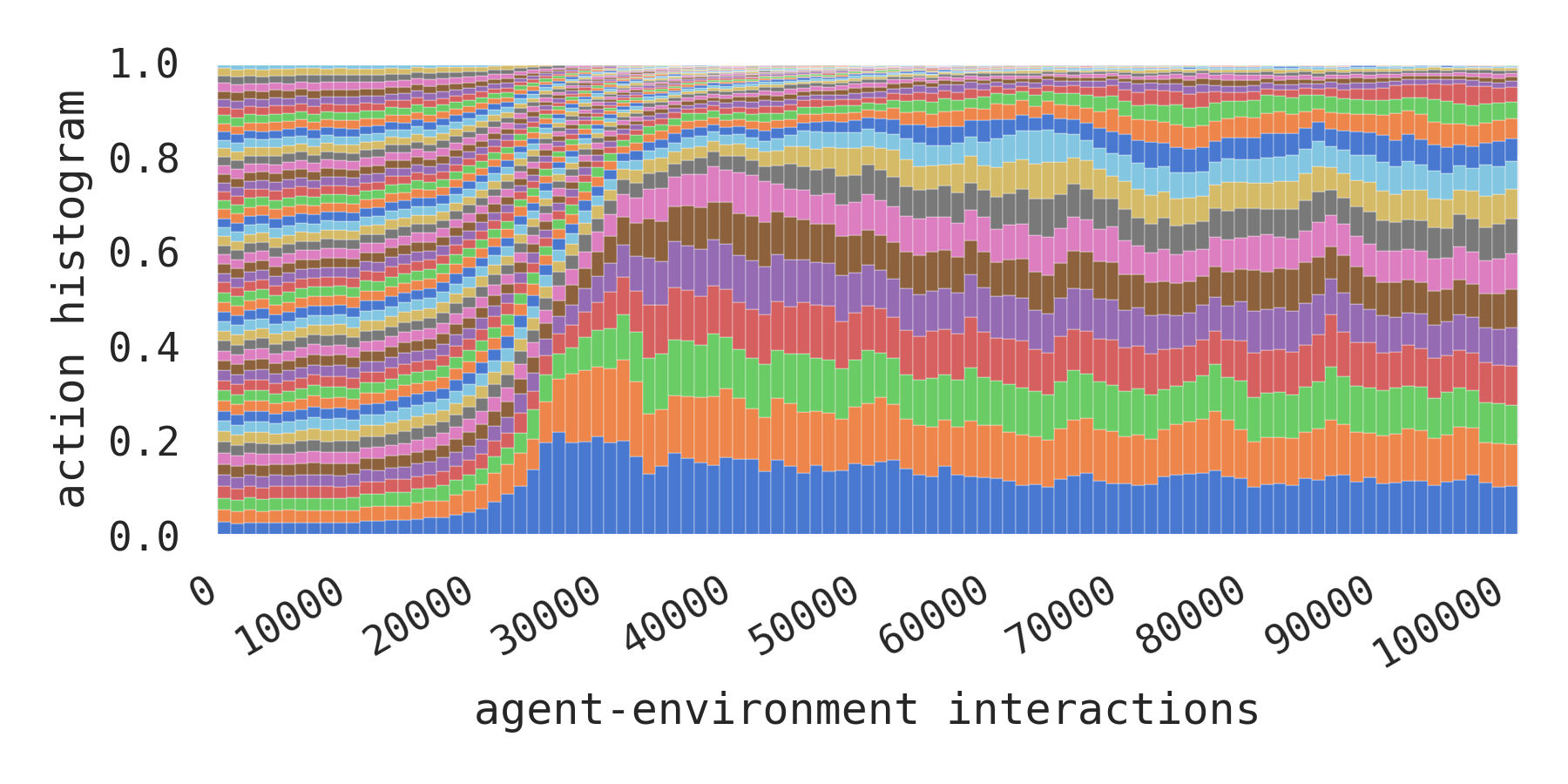}
        \caption{MMD regularization}
    \end{subfigure}
    \\
    \begin{subfigure}{.32\linewidth}
        \includegraphics[width=\linewidth]{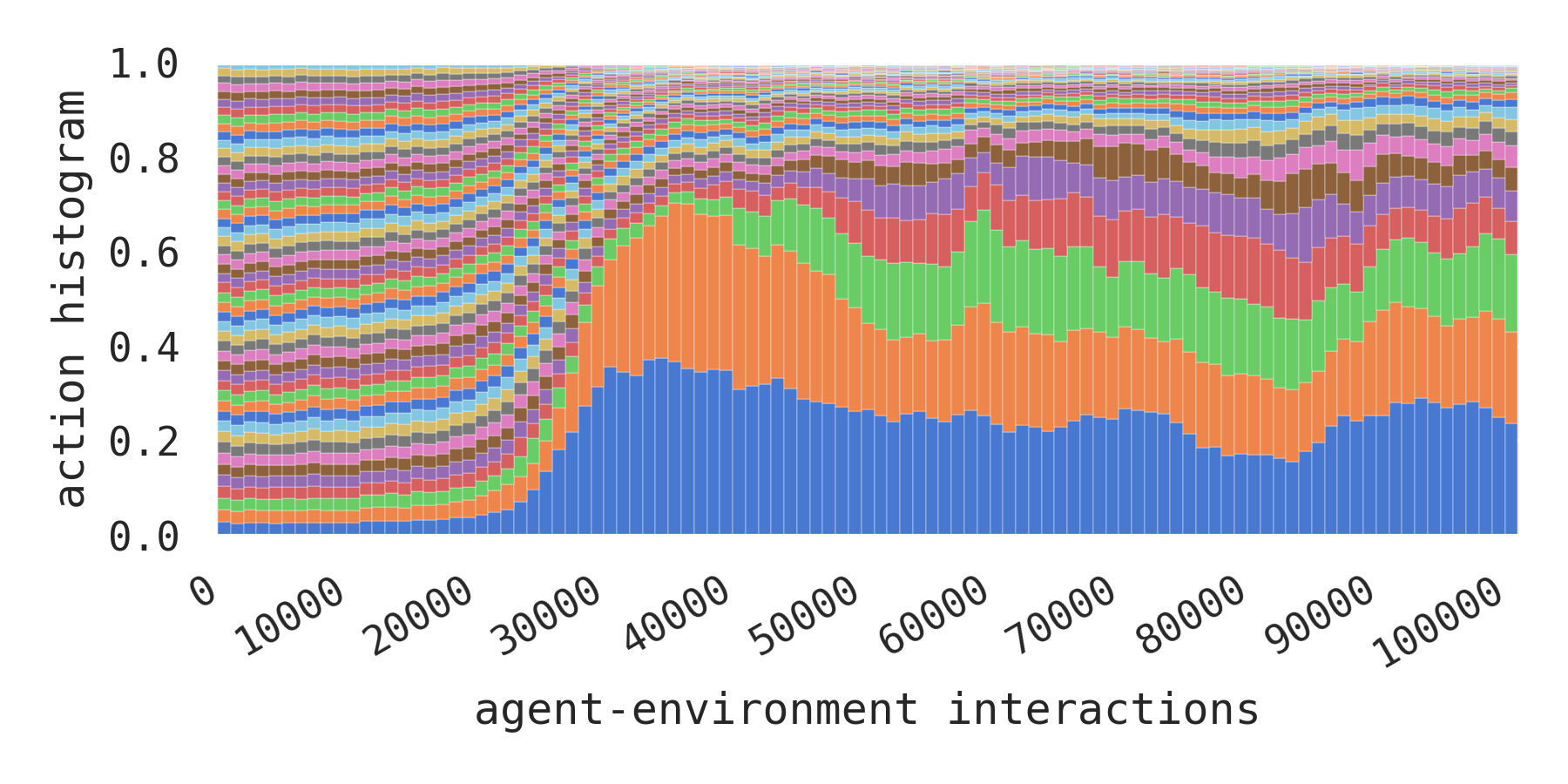}
        \caption{Jensen-Shannon regularization}
    \end{subfigure}
    \begin{subfigure}{.32\linewidth}
        \includegraphics[width=\linewidth]{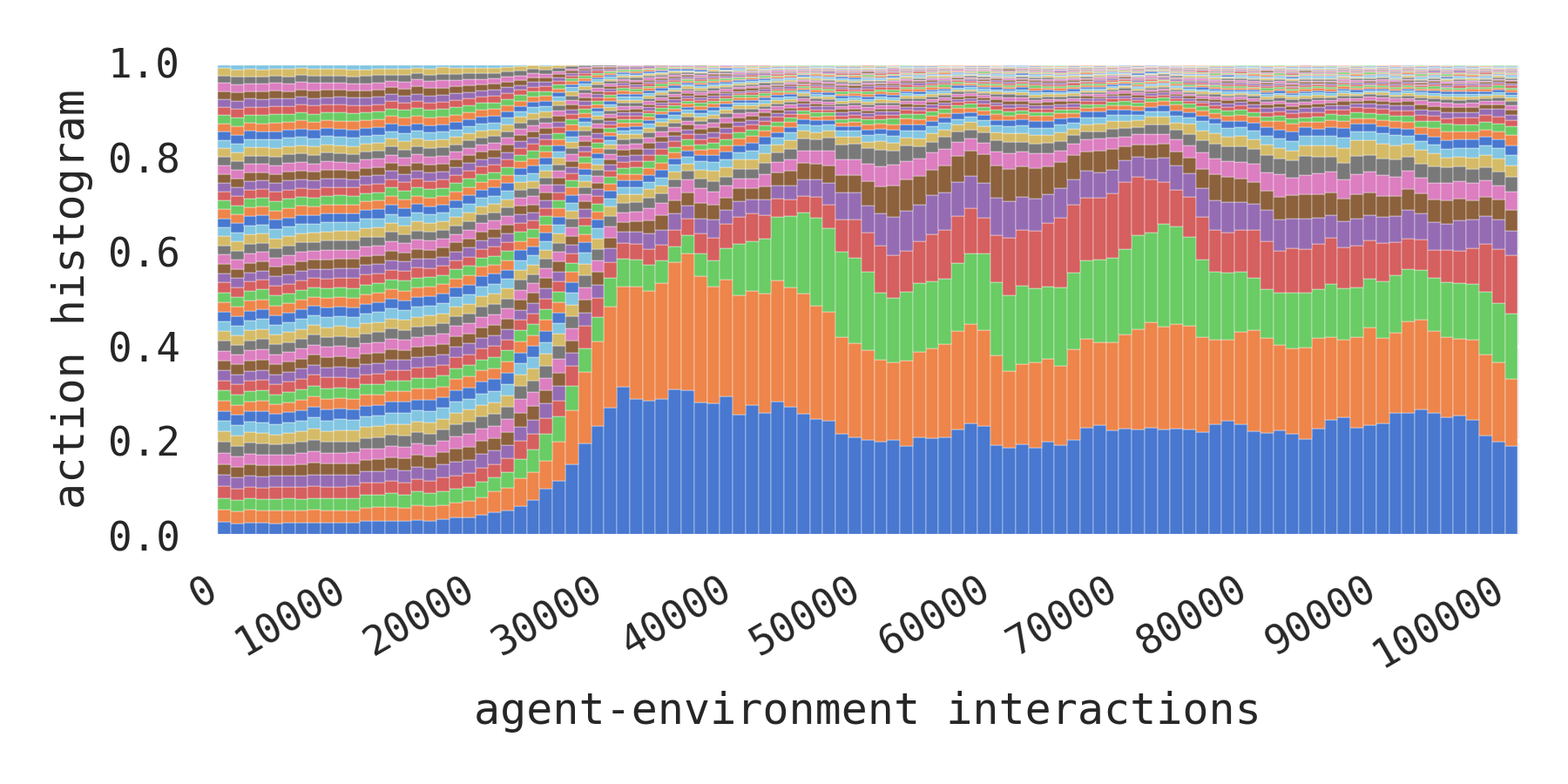}
        \caption{Hellinger regularization}
    \end{subfigure}
    \begin{subfigure}{.32\linewidth}
        \includegraphics[width=\linewidth]{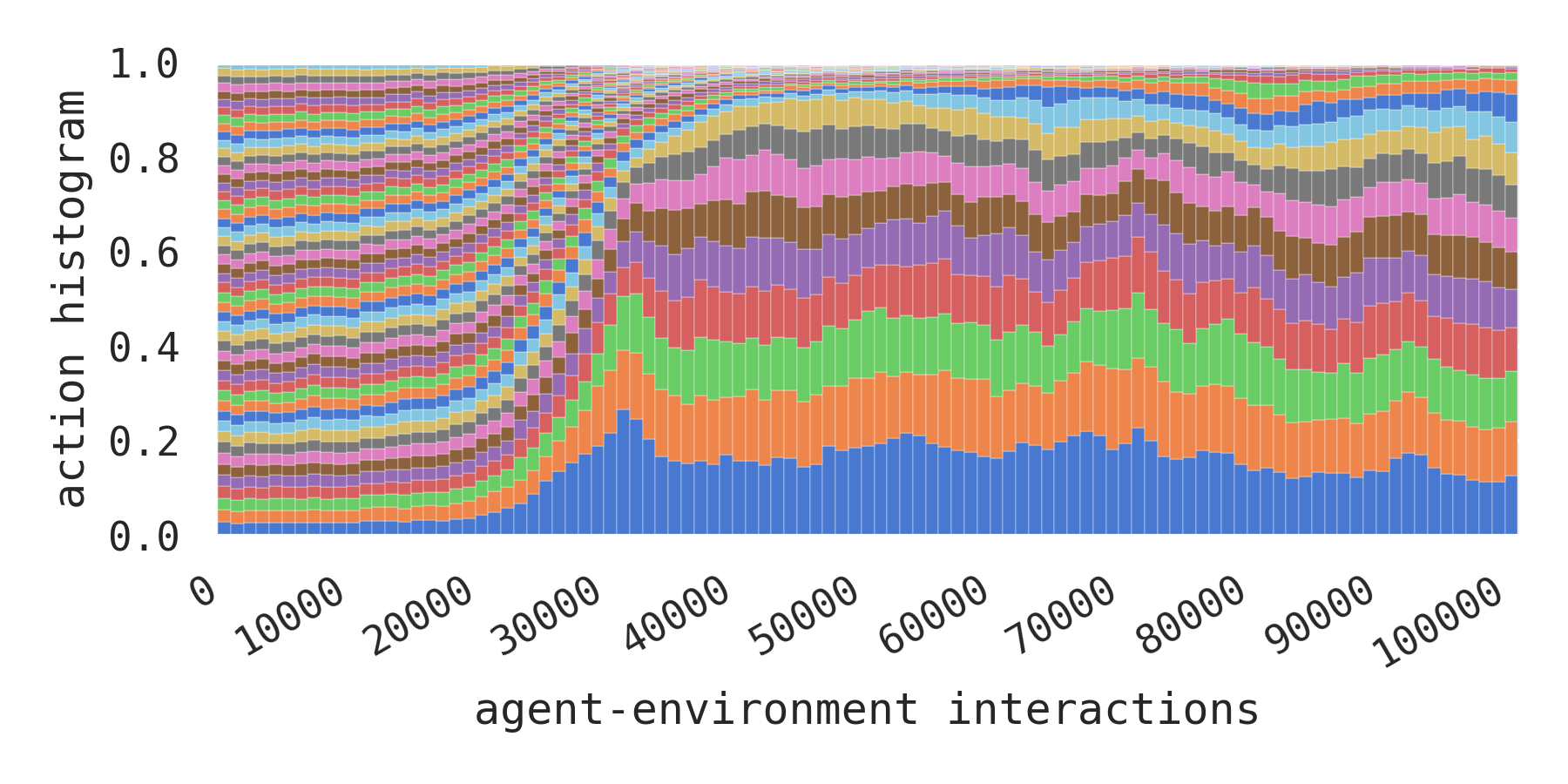}
        \caption{Total variation regularization}
    \end{subfigure}
    \caption{Results of image classification experiment on Spotify environment.}
    \label{fig:spotify}
\end{figure*}

In this experiment we set up a synthetic music recommendation system proposed in~\cite{Dereventsov_2022}.
We use Spotify Web API\footnote{\url{https://developer.spotify.com/documentation/web-api/}} to construct a contextual bandit environment that replicates the task of track recommendation to a user.
In this setting the observations (users) are synthetically generated and are represented by their preferences to various musical genres, and the actions are given by the set of tracks the agent can recommend.
The reward for recommending a track to a user is either $1, -1,$ or $0$, indicating that the user liked/disliked/did not provide feedback, respectively.
See~\cite{Dereventsov_2022} for a more detailed explanation of the environment.
The agent reward, policy entropy, and action selection histograms for the various regularizers on Spotify environment are shown in Figure~\ref{fig:spotify}.

We note that while all agents achieve satisfactory performance in terms of reward, the action selection of the baseline agent is constrained to only $3$ tracks (out of $50$ possible), which is neither practical nor acceptable in real-world applications.
All regularized agents provide a much more diverse action selection, while also achieving higher reward values.

A particular interest of this environment is the fact that there are infinitely many policies that achieve near-optimal performance.
As an example, even the unregularized policy gradient agent almost learned the environment, while only ever taking about $6\%$ of the available actions, with one action being taken about $40\%$ of the time.
In comparison, for regularized agents the action selection is much more diverse with fewer ``favorite'' actions.
Most notably, the MMD-regularized agent is actively taking about $40\%$ of the actions with the most frequent one being selected only about $8\%$ of the time.

%%%%%=====================
\section{Concluding remarks}
\label{sec:conclusions}
%%%%%=====================

In this effort, we consider the impact of regularization on the diversity of actions taken by policies generated from policy gradient RL agents. 
In the context of personalized RL there are several addition advantages that extend from this work.  First, the $\varphi$-divergence and MMD-based regularization encourages exploration and aids to  prevent early convergence to sub-optimal policies. Second, the resulting policies can serve as a good (macro) initialization for a more (micro) specific behavior. Finally, the resulting policies are more robust in the face of adversarial perturbations or noise as evidenced by our various numerical examples.  

However, there is much more extensive testing to be done and a supporting theory needs to be developed before any victories can be declared.  As mentioned throughout, there has been extensive amounts of research by the RL community on using KL-type entropy regularization, but more advanced discrepancies such as the MMD-based approach we presented here are still in their infancy. 
In addition, there is a vast amount of research on optimal transport theory which, in connection with entropy-type penalization is something we also plan to investigate.  These methods possess some computational challenges but have the ability to lift a ground metric from the data-space to the set of probability measures on this space and, therefore,  take into account the underlying geometry of the data 
\cite{Kantorovich1942Russian,Kantorovich2006OnTT, NIPS2006_e9fb2eda}. To our knowledge, this area of research has yet to be explored by the machine learning community. 

%%%%%=====================
%\section*{Acknowledgment}
%\label{sec:ack}
%%%%%=====================

%%%%%=====================
%References
%%%%%=====================
\bibliography{RL,clay}

\end{document}